\documentclass[review]{elsarticle}
\usepackage{hyperref}
\usepackage{algorithm}
\usepackage{amsmath}
\usepackage{amssymb}
\usepackage{graphicx}
\usepackage{subfigure}
\usepackage{color}
\usepackage{float}
\usepackage{sidecap}
\usepackage{rotating}
\usepackage{booktabs}
\usepackage{multirow}
\usepackage{ulem}
\usepackage{algpseudocode}

\begin{document}

\begin{frontmatter}

\title{Learning to Fuse Local Geometric Features for 3D Rigid Data Matching}
\author{Jiaqi~Yang}
\ead{jqyang@hust.edu.cn}

\author{Chen Zhao}
\ead{hust\_zhao@hust.edu.cn}

\author{Ke~Xian}
\ead{kexian@hust.edu.cn}

\author{Angfan~Zhu}
\ead{zhuangfan@hust.edu.cn}

\author{{Zhiguo Cao}\corref{corre}}
\ead{zgcao@hust.edu.cn}
\cortext[corre]{Corresponding author.}

\address{National Key Laboratory of Science and Technology on Multi-spectral Information Processing, School of Artificial Intelligence and Automation, Huazhong University of Science and Technology, P. R. China}

\begin{abstract}
This paper presents a simple yet very effective data-driven approach to fuse both low-level and high-level local geometric features for 3D rigid data matching. It is a common practice to generate distinctive geometric descriptors by fusing low-level features from various viewpoints or subspaces, or enhance geometric feature matching by leveraging multiple high-level features. In prior works, they are typically performed via linear operations such as concatenation and min pooling. We show that more compact and distinctive representations can be achieved by optimizing a neural network (NN)  model under the triplet framework that  non-linearly fuses local geometric features in Euclidean spaces. The NN model is trained by an improved triplet loss function that fully leverages  all pairwise relationships within the triplet. Moreover, the fused descriptor by our approach is also competitive to deep learned descriptors from raw data while being more lightweight and  rotational invariant. Experimental results on four standard datasets with various data modalities and application contexts confirm the advantages of our approach in terms of both feature matching and geometric registration.
\end{abstract}
\begin{keyword}
Local geometric feature\sep feature fusion \sep deep learning\sep feature matching\sep point cloud registration
\end{keyword}
\end{frontmatter}
\section{Introduction}
Establishing reliable point-to-point correspondences between 3D rigid shapes represented by point clouds, meshes, or depth maps is a fundamental issue in 3D computer vision, computer graphics, photogrammetry, and robotics. To judge whether two points are corresponding, a typical solution is first representing the geometric information of the proximity of a point by a feature vector, known as the local geometric descriptor, and then measuring the similarity between their feature descriptors. As such, distinctive and robust feature representations are desired to achieve reliable shape correspondences.

Following the trend in 2D vision area~\cite{lowe2004distinctive,Yu2015Multi}, a number of 3D local geometric descriptors, either hand-crafted~\cite{johnson1999using,rusu2009fast,yang2017RCS_jrnl,tombari2010unique,guo2013rotational,yang2016fast,yang2017toldi} or learned~\cite{zeng20173dmatch,khoury2017learning,elbaz20173d,deng2018ppfnet,yew20183dfeat,Deng2018PPFold}, have been proposed for the purpose of fully encoding the geometric information of a rigid local shape (this paper concerns about \textit{rigid} data only, some literature about non-rigid data feature representations will be introduced in Sect.~\ref{sec:related}). We can find a variety of low-level geometric features employed to represent the local 3D structure, e.g., normal deviation~\cite{rusu2009fast}, signed distance~\cite{malassiotis2007snapshots}, contour~\cite{yang2017RCS_jrnl}, and density~\cite{guo2013rotational}. To achieve information complementary, many descriptors suggest aggregating multiple low-level features from different subspaces~\cite{tombari2010unique,yang2016fast} or viewpoints~\cite{guo2013rotational,yang2017RCS_jrnl}, which has been demonstrated to be a simple yet effective way~\cite{guo2013rotational,yang2017RCS_jrnl}. These descriptors mainly resort to concatenation for feature fusion. However, it has two key drawbacks. One is the information redundancy and resultant high-dimensional feature vectors that are inefficient to store and match.  The other is that all bins in the considered low-level features are enforced with identical weights, which appears to be improper because each bin may contribute differently (even resulting in negative impacts) in a specific scenario~\cite{yang2016accumulated}.  There are also some works that perform Principle Component Analysis (PCA) on the concatenated feature, e.g.,~\cite{johnson1999using,guo2015novel}. Although more compact feature variants can be achieved with PCA, PCA still fails to discover a reasonable feature embedding to integrate the beneficial information provided by each feature bin while suppressing the impact of noisy bins simultaneously~\cite{khoury2017learning}, possibly owing to the lack of matching supervision.

In addition to low-level feature fusion, some studies in the image matching realm suggest using multiple feature descriptors to perform feature matching~\cite{hsu2015robust,hu2016progressive} because the optimal descriptor for feature matching may vary from image to image, or even pixel to pixel. This idea has later been applied to 3D feature matching in~\cite{buch2016local} where a min pooling method is proposed to fuse high-level geometric features. Above fusion methods for high-level features tend to select the optimal feature for a particular pair of image/surface patches. Accordingly, those \textit{sub-optimal} features are discarded to measure the similarity of two patches. However, we will show that there still exists valuable information in other features that contributes to more accurate patch similarity calculation.

Although the learning for local geometric feature fusion remains unexplored in the field of 3D rigid data matching, there are already a number of deep learning-based local geometric descriptors~\cite{zeng20173dmatch,khoury2017learning,deng2018ppfnet} and the majority of them directly learn features from raw data. For the task of scene registration, these learned descriptors have already outperformed traditional descriptors by a clear margin. Unfortunately, most of existing learned descriptors are intrinsically sensitive to rotation~\cite{Charles2017PointNet}, which greatly limits their deployment in real-world applications. By contrast, the fusion result by learning from a set of rotational invariant traditional feature representations, as will be done in this work, inherits the merit of being invariant to rotation~\cite{khoury2017learning}. Also, we will show that an ultra lightweight network is  sufficient to learn a reasonable feature fusion and the resultant descriptor is competitive to existing deep learned descriptors.
\begin{figure}[t]
	\centering
	\includegraphics[width=0.7\linewidth]{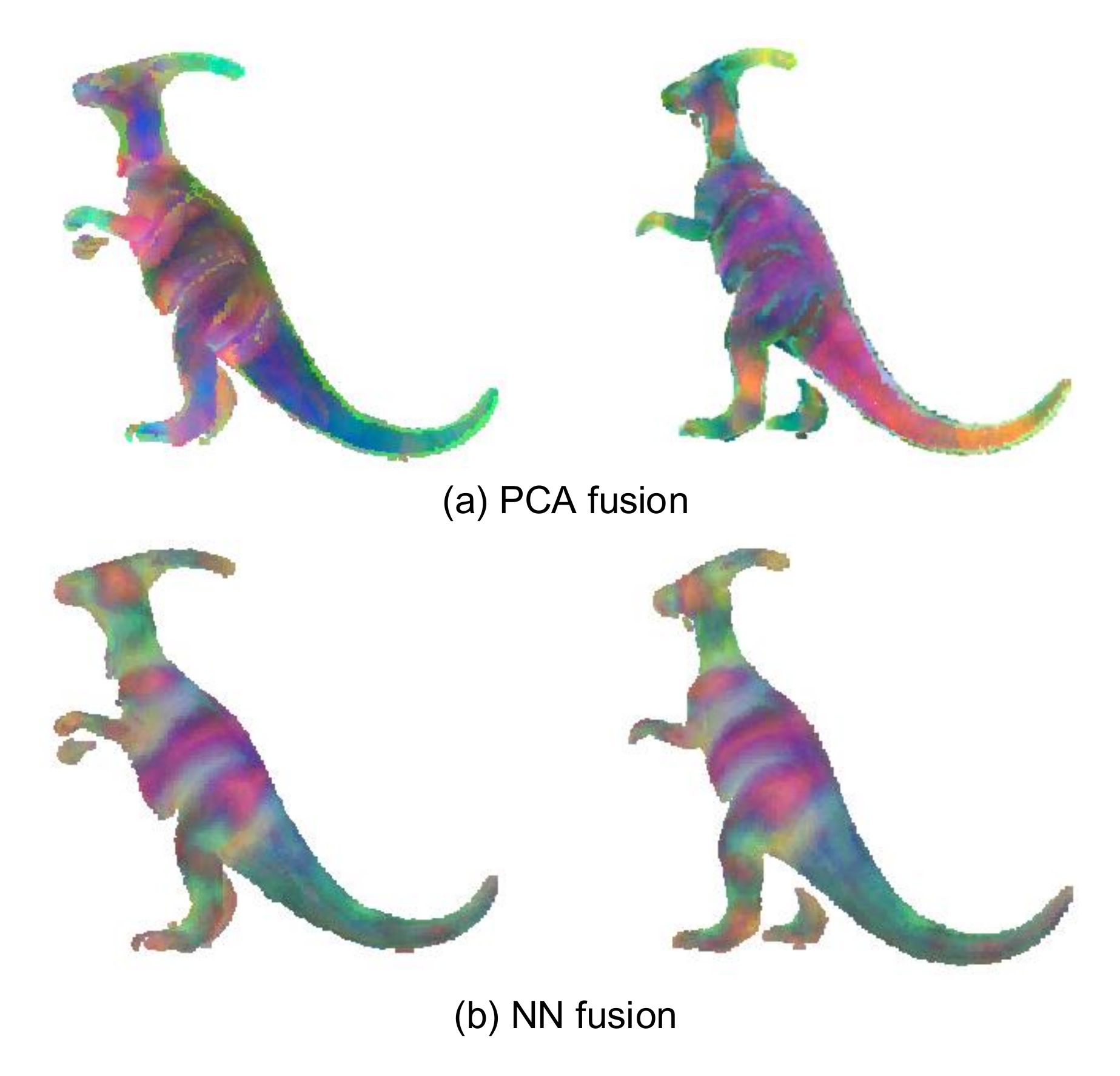}\\
	\caption{Visualization of fused features (features were mapped into the RGB space) of SHOT~\cite{tombari2010unique}, RoPS~\cite{guo2013rotational}, and RCS~\cite{yang2017RCS_jrnl} over two views of the \textit{parasaurolophus} model~\cite{mian2006novel}. (a) Result of PCA that projects the concatenated feature vector of SHOT, RoPS, and RCS into a low-dimensional space. (b) Result of our method, which leverages a non-linear feature fusion scheme based on a neural network (NN) to fully combine the geometric information provided by different features. The dimensionality of fused features by PCA and NN is 256.}
	\label{fig:1}
\end{figure}

Motivated by these considerations, we present a deep learning-based approach to fuse local geometric features for 3D rigid data matching. More specifically, a compact and distinctive new feature is generated by feeding multiple low/high-level features to a  Neural Network (NN) under the triplet  framework. (i) Compared with previous  linear fusion methods,  we show that NN owes the ability to discover and fully leverage the advantageous information provided by each feature with a non-linear feature embedding. A comparison of our NN-fused feature and PCA-fused feature can be found in Fig.~\ref{fig:1}. (ii) Compared with existing deep learned descriptors, our method is fully rotational invariant while requiring a dramatically more lightweight network for training. We show that our method promotes the distinctiveness of traditional local geometric feature descriptors greatly via learning for fusion, achieving competitive performance with several deep learned descriptors on a scene registration benchmark. To train the network, we propose a new loss that improves the original triplet loss from two aspects. First, we consider all negative pairs in the triplet to compute the negative distance to more reliably separate matches and non-matches. Second, a pairwise term for pulling the positive samples together is attached to achieve a tighter cluster formed by matches. Experiments have been carried out on four standard datasets with different application contexts and data modalities. Comparison with prior geometric fusion methods and several deep learned descriptors confirms the overall superiority of the presented approach. Our main contributions are summarized as follows:
\begin{itemize}
	\item We present a deep learning-based approach to non-linearly fusing multiple local geometric features and reveal that neural networks owe the ability to effectively discover and integrate the complementary information within a geometric feature set.
	\item An improved triplet loss that explores all pairwise relationships within the triplet  is proposed. It compares favorably to the classical triplet loss and contrastive loss in the local geometric feature fusion context.
	\item Compared with deep learned local geometric descriptors, the descriptor produced by the proposed fusing approach achieves comparable performance, albeit being more lightweight to train and truly rotational invariant. Fusing traditional local geometric features with deep learning methods therefore paves a new path for the research of local geometric feature description.
	
\end{itemize}

The remainder of this paper is organized as follows. Sect.~\ref{sec:related} gives a review of relevant literature for local geometric feature fusion. Sect.~\ref{sec:mtd} details the proposed fusion method using a triplet network architecture for both low-level and high-level local geometric features. Sect.~\ref{sec:exp} presents the experiments deployed to validate the effectiveness of our method together with necessary explanations and discussions. Sect.~\ref{sec:conc} draws the conclusions and discusses about potential future research directions.

\section{Related work}\label{sec:related}

\noindent\textbf{Low-level geometric feature fusion.} Fusion of several low-level geometric features from multiple viewpoints or subspaces is prevalent in modern local geometric descriptors. 3D-to-2D projection provides a straightforward way to describe the local geometry by virtue of 2D cues. Representative examples include Point Fingerprint~\cite{sun2003point} and Snapshots~\cite{malassiotis2007snapshots}. To gain more discriminative information, the \textit{rotation and projection} mechanism is proposed in Rotational Projection Statistics (RoPS)~\cite{guo2013rotational} and Rotational Contour Signatures (RCS)~\cite{yang2017RCS_jrnl} that concatenate features extracted from multiple viewpoints. There also exist methods that perform feature description in the 3D space such as 3D Shape Context~\cite{frome2004recognizing} and Signatures of Histograms of OrienTations (SHOT)~\cite{tombari2010unique}, usually with subspace partition to achieve spatial information characterization. In these descriptors, features from different spatial subspaces are concatenated as the final descriptor. With the observation that different geometric attributes could bring complementary information to each other, Local Feature Statistics Histograms~\cite{yang2016fast} concatenates histograms of normal deviation, signed distance, and point density. Besides concatenation, Triple Spin Image (TriSI)~\cite{guo2015novel} applies PCA to the concatenated feature vector consists of three spin image signatures to achieve a more compact variant. Either concatenation or PCA is employed for  descriptors based on low-level geometric feature fusion in existing literature. 
\\
\\
\noindent\textbf{High-level geometric feature fusion.} (i) \textit{Non-rigid shape  retrieval.} A typical application of the fusion of multiple high-level geometric features can be found in 3D non-rigid object retrieval studies. For non-rigid data, one of the most challenging factors is  deformation~\cite{Tam2013Registration}. The fusion of multiple or multi-scale high-level features such as Heat Kernel Signatures (HKS)~\cite{sun2009concise} and Wave Kernel Signatures (WKS)~\cite{aubry2011wave} is a popular way for improving the performance of  shape retrieval.  Yang and Chen~\cite{yang2007ofs} presented an optimized feature selection strategy for combining a set of high-level features for the task of object retrieval.  Papadakis et al.~\cite{papadakis20083d} concatenated multiple 3D features based on spherical harmonics as well as 2D features based on depth buffers as the feature vector, which is further compressed via scalar quantization. Tabia et al.~\cite{tabia2014covariance} proposed using the covariance matrices of the descriptors to efficient fuse shape descriptors for 3D face matching and retrieval. Xie et al.~\cite{xie2015deepshape} proposed a deep shape descriptor for 3D object retrieval that concatenates the features from the hidden layers of several auto-encoder networks. Fang et al.~\cite{Yi20153D}  proposed combining HKS and heat shape descriptor  to parametrize raw 3D shapes for deep feature learning. Bu et al.~\cite{Bu2014Multimodal} proposed fusing the different modalities of 3D shapes via deep learning to promote the discriminability of unimodal feature.  Furuya et al.~\cite{Furuya2016bmvc} proposed a deep local feature aggregation network that integrates the extraction of rotation-invariant 3D local features and their fusion in a single deep network, they finally aggregated these local features to a global feature for shape retrieval. 
(ii) \textit{Rigid 3D matching.}  In the field of 3D rigid data matching, min pooling proposed in~\cite{buch2016local}, to the best of our knowledge, is the only approach that leverages multiple 3D feature descriptors to enhance the feature matching performance in 3D object recognition context. Specifically, min pooling suggests selecting the feature resulting in the highest feature similarity to judge two patches as a match or non-match. However, potential complementary information of other descriptors are therewith discarded.
\\
\\
\noindent\textbf{Learning for feature fusion.} Since  the deep learning architecture provides flexible interfaces to tasks with various inputs and allows outputting aggregated results such as classification vectors and regression values from multiple input channels, it is straightforward to fuse features within a deep learning framework. For instance, deep fusion of RGB and depth features~\cite{gupta2014learning}, view features from different viewpoints~\cite{su2015multi}, and different geometric features~\cite{fang20153d}. Besides deep learning, there is also a field called multi-view learning that introduces one function to model a particular feature and jointly optimizes all the functions to find the redundant features of the same input data as well as improve the learning performance~\cite{xu2013survey,Jing2017Multi}. It includes three categories of learning approaches, i.e., co-training, multiple kernel learning, and subspace learning. They have been successfully applied to a number of computer vision tasks such as object classification~\cite{kumar2007support}, object recognition~\cite{kembhavi2009incremental}, facial expression recognition~\cite{zheng2006facial}, image classification~\cite{zhang2012combining}, and image retrieval~\cite{li2011difficulty}. Nonetheless, above learning-based feature fusion techniques have not found applications in the realm of 3D rigid shape matching yet. Inspired by the great success of deep learning for feature fusion in previous mentioned applications, we present the first attempt to fuse local geometric features to improve feature matching and rigid geometric registration with a deep neural network.
\section{Method}\label{sec:mtd}
In this section, we detail the proposed fusion method for local geometric features. More specifically, we train a deep neural network to combine a number of local geometric features $\{{\bf f}_i\}_{i=1,\cdots,N}$ to a new feature ${\bf f}_{nn}$ that achieves information complementary while being compact as well.
\subsection{Network architecture}
\begin{figure}[t]
	\centering
	\includegraphics[width=1.0\linewidth]{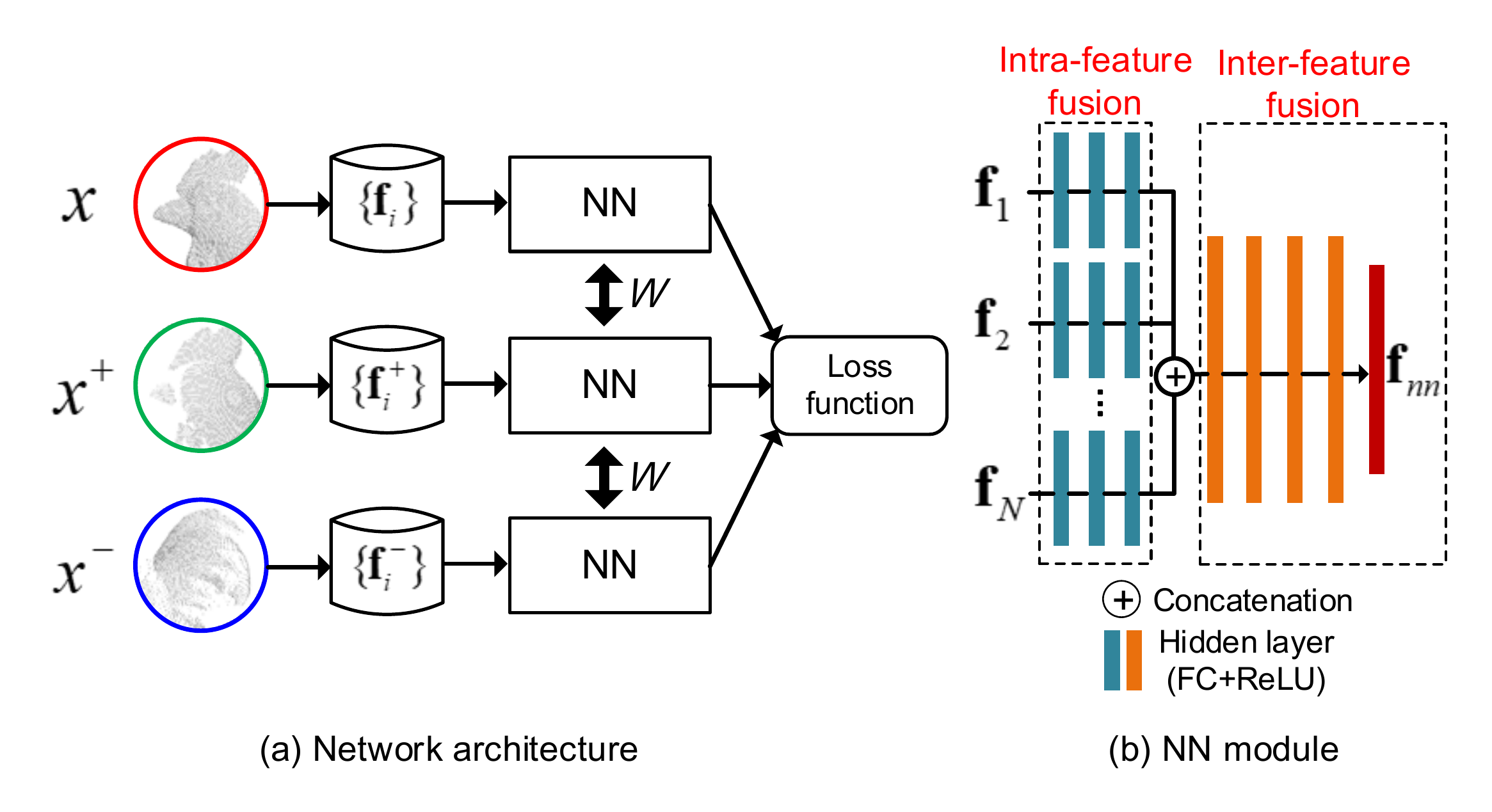}\\
	\caption{Architecture of the proposed network for local geometric feature fusion. (a) The triplet framework with shared weights that simultaneously takes three samples, i.e., an anchor sample $x$, a positive sample $x^+$, and a negative sample $x^-$, as the input. For each sample, a set of geometric features $\{{\bf f}_i\}$ are extracted and then fed to an NN module that merges these features as a new feature. The new feature is then used to compute the network loss for parameters updating. (b) Structure of the NN module. NN consists of an intra-feature fusion block and an inter-feature fusion block. The intra-feature fusion block for each feature includes 3 hidden layers. Features after intra-feature fusion processing are then concatenated and passed to the inter-feature fusion block with 4 hidden layers and 1 output layer (i.e., the eventual fused feature by the network). Each hidden layer consists of a fully connected layer and an ReLU layer.}
	\label{fig:network}
\end{figure}

Learned descriptors in 2D and 3D domains have been typically trained with 2-branch~\cite{simo2015discriminative,zeng20173dmatch}, 3-branch~\cite{kumar2016learning,khoury2017learning}, and N-branch~\cite{deng2018ppfnet} Siamese networks. The 3-branch network is shown to be more suitable than the 2-branch network with feature histograms being the input~\cite{khoury2017learning}, and requires far less hardware resource than the N-branch network. Therefore, we adopt the Siamese network under the triplet frame as our basic architecture with the purpose of pulling matches together while pushing non-matches apart. 

Fig.~\ref{fig:network} presents the proposed network architecture. Let $x$, $x^+$, and $x^-$ respectively be an anchor sample, a positive sample, and a negative sample that will be fed to the network together. In our context, the training samples refer to local shape patches. We first extract the geometric features to be fused for each sample. Then, the features of each sample are embedded by an NN that non-linearly merges all input features as a  new feature in Euclidean spaces. We keep the weights of the three NN module identical during parameter updating. For the NN module, two blocks are included, i.e., an intra-feature fusion block and an inter-feature fusion block. The intra-feature fusion block allows bin-level fusion for each independent feature. Since the bins of the feature after intra-feature fusion are combinations of all initial feature bins with different weights, this can be seen as information exchange within a feature. By contrast, the most popular concatenation operation for local geometric fusion  sets identical weights to all bins within a feature and ignores the combinational information of bins. The features after intra-feature fusion are then concatenated as a single vector, and finally passed to the inter-feature fusion block that combines feature-level information. 

We use 3 fully connected (FC) layers in the intra-feature fusion block for each feature respectively with $N_{intra}$, $N_{intra}$, and $\frac{N_{intra}}{2}$ nodes, where each FC layer is followed by an ReLU layer.  The inter-feature fusion block is composed of 5 FC + ReLU layers where each of the former 4 FC layers has $N_{inter}$ nodes and the last FC layer (i.e., the final fused feature) has $N_{nn}$ nodes. The values of $N_{intra}$ and $N_{inter}$ will be tuned for a specific input feature combination. We will also consider different values of  $N_{nn}$ to achieve a balance between compactness and distinctiveness.
\subsection{Loss function}
The network is trained to minimize an improved triplet loss proposed in this paper. We  first revisit the original triplet loss in the following. 

Let $(\cal F, {\cal F}^+, {\cal F}^-)$ be the extracted feature sets for samples $(x, x^+, x^-)$ where ${\cal F}=\{{\bf f}_i\}_{i=1,\cdots,N}$ denotes the $N$ local geometric features to be fused. Consider a set of triplets of $\cal F$, i.e., ${\cal S}=\{({\cal F}_j, {\cal F}_j^+, {\cal F}_j^-)\}_j$. The triplet loss is defined as:
\begin{equation}\label{eq:tripletloss}
{\cal L}({\boldsymbol \theta}) =\frac{1}{|\cal S|}\sum\limits_{j = 1}^{|\cal S|}[d(f({\cal F}_j;{\boldsymbol \theta}), f({\cal F}^+_j ;{\boldsymbol \theta}))-d(f({\cal F}_j;{\boldsymbol \theta}), f({\cal F}^-_j ;{\boldsymbol \theta}))+ \tau_{tri}]_+,
\end{equation}
where $d(\cdot,\cdot)$ denotes the $L_2$ distance between two vectors, $\bf{\theta}$ are the parameters of  mapping $f$,   $[]_+$ denotes $\rm{max}(\cdot,0)$, and $\tau_{tri}$ is the margin set to separate positive and negative samples. 

\begin{figure}[t]
	\centering
	\includegraphics[width=1.0\linewidth]{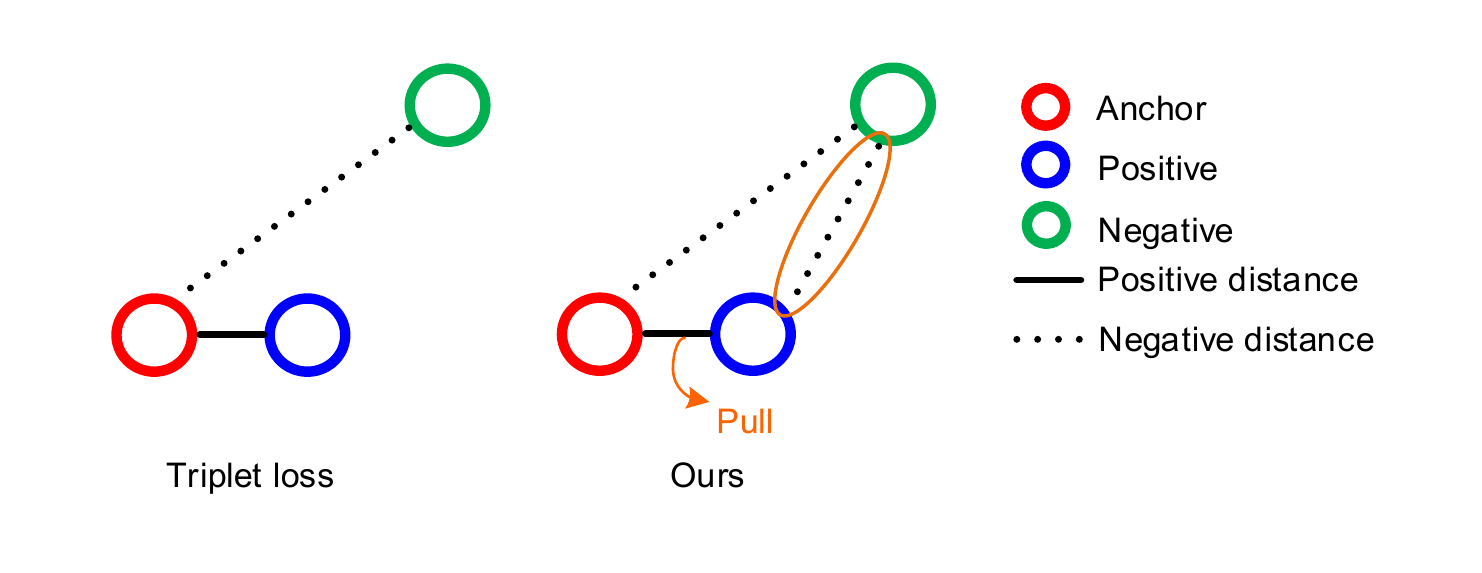}\\
	\caption{Comparison between the triplet loss and our improved loss. In our loss, we consider both negative distances to better separate positive and negative samples. In addition, we pull the anchor and positive samples to help the network form a tighter  positive sample cluster.}
	\label{fig:loss}
\end{figure}
However, the original triplet loss fails to fully exploit the relationships within the triplet from two aspects. First, there are two negative distances, i.e., $d(f({\cal F}_j;{\boldsymbol \theta}), f({\cal F}^-_j ;{\boldsymbol \theta}))$ and $d(f({\cal F}_j^+;{\boldsymbol \theta}), f({\cal F}^-_j ;{\boldsymbol \theta}))$, while it only considers the former one. Since the objective of the triplet loss is separating positive and negative samples by a predefined margin, it is more reasonable to consider all negative distances. Second, the triplet loss ignores the compactness of the cluster formed by positive samples. Intuitively, negative samples may distribute scatteredly in the feature space because they are supposed to be distant from the anchor sample and no pairwise affinity exists between two negative samples. By contrast, positive samples show clear consistency as the distance between any two positive samples should close to 0 in the ideal case. In the context of geometric feature matching where  perturbations such as noise, partial overlap, clutter, and occlusion commonly exist in real-world applications, even two corresponding keypoints may exhibit dissimilar geometric features. Therefore, it is desired to further pull the anchor and positive samples such that challenging matches typically located in boarder or incomplete regions can be recognized by the network during testing. 

Under these concerns, we propose the following improved triplet loss:
\begin{equation}\label{eq:loss}
\widehat{\cal L}({\boldsymbol \theta}) =\frac{1}{|\cal S|}\sum\limits_{j = 1}^{|\cal S|}[d(f({\cal F}_j;{\boldsymbol \theta}), f({\cal F}^+_j ;{\boldsymbol \theta}))-d^*+ \tau_{tri}+{\tau_{pair}}d(f({\cal F}_j;{\boldsymbol \theta}), f({\cal F}^+_j ;{\boldsymbol \theta}))]_+,
\end{equation}
where $d^*$ is the minimum of $d(f({\cal F}_j;{\boldsymbol \theta}), f({\cal F}^-_j ;{\boldsymbol \theta}))$ and $d(f({\cal F}_j^+;{\boldsymbol \theta}), f({\cal F}^-_j ;{\boldsymbol \theta}))$, and $\tau_{pair}$ is a parameter to control the weight of the last pairwise term. Compared with the triplet loss (Eq.~\ref{eq:tripletloss}), our improved loss considers both negative distances within the triplet to ensure a better separation for positive and negative samples, and minimizes the positive distance to achieve a tight positive sample cluster. 
\subsection{Training data preparation}\label{subsec:data}
\begin{figure}[t]
	\centering
	\includegraphics[width=0.9\linewidth]{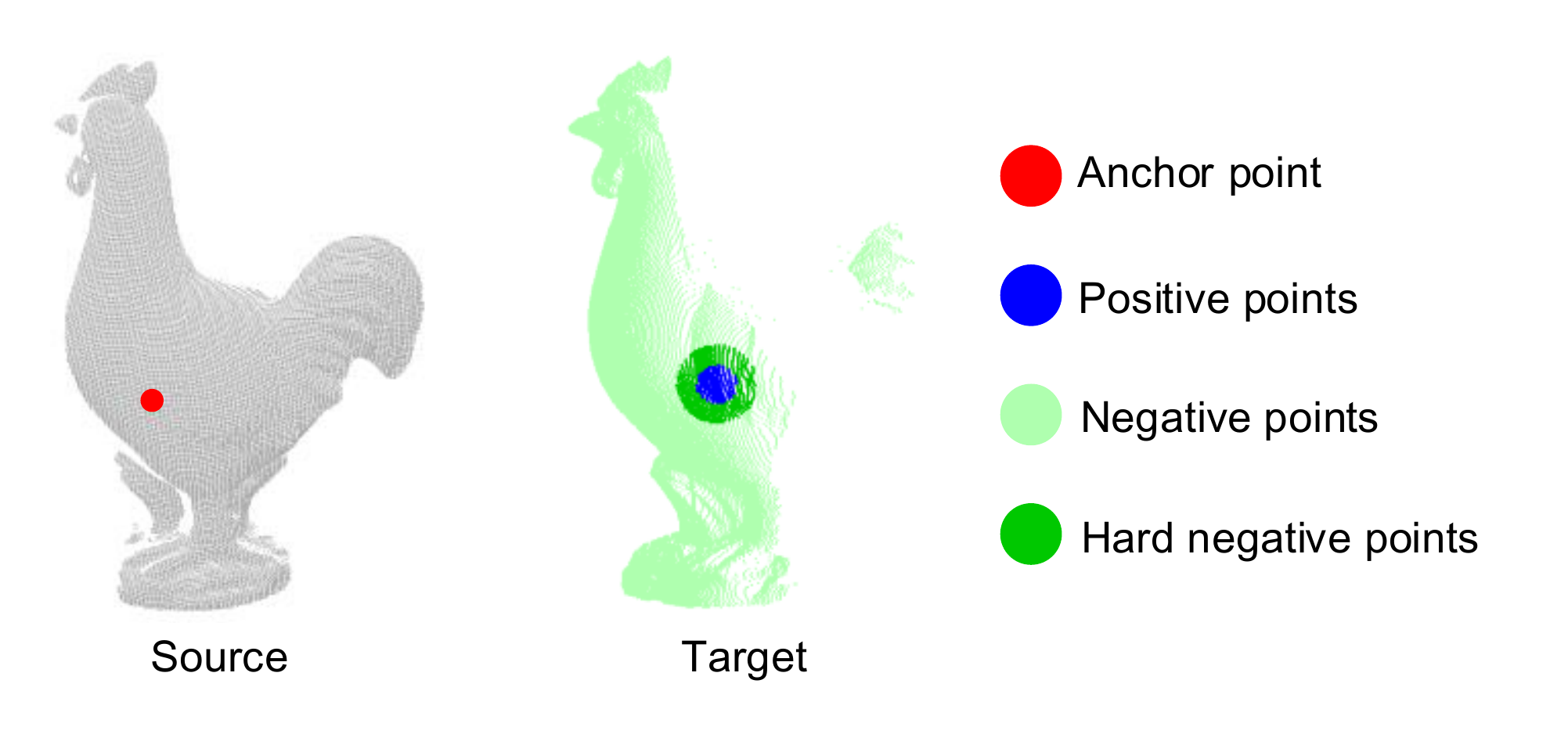}\\
	\caption{Exemplar illustration of the sampling of triplets for training over two partially overlapped shapes. }
	\label{fig:sampling}
\end{figure}
We sample triplets of points for feature extraction and training from a number of shape pairs with overlaps. Given a shape pair composed of a source shape ${\cal P}^s$ and a target shape ${\cal P}^t$ together with the ground truth rotation ${\bf R}_{GT}$ and translation ${\bf t}_{GT}$, any point in ${\cal P}^s$, e.g., ${\bf p}^s_k$, that satisfies the following condition will be considered as an anchor sample:
\begin{equation}
||{\hat{\bf p}}^s_k-{\rm nn}({\hat{\bf p}}^s_k, {\cal P}^t)||_2<\tau_{anc},
\end{equation}
where ${\hat{\bf p}}^s_k={\bf p}^s_k{\bf R}_{GT}+{\bf t}_{GT}$, ${\rm nn}({\hat{\bf p}}^s_k, {\cal P}^t)$ is the nearest neighbor of ${\hat{\bf p}}^s_k$ in the target shape ${\cal P}^t$, and $\tau_{anc}$ is a distance threshold. To select the positive samples for ${\bf p}^s_k$, we find points in ${\cal P}^t$ whose distances to ${\hat{\bf p}}^s_k$ are smaller than $\tau_{+}$ and serve them as the positive samples to ${\bf p}^s_k$. As for negative samples, we follow~\cite{khoury2017learning} that divides the negative examples to be sampled into two portions. The first portion includes $N^{hard-}_k$ samples randomly selected from the set of points in ${\cal P}^t$ that are at distance at least $\tau_{+}$ and at most $\tau_{hard-}$ from ${\hat{\bf p}}^s_k$. Because these points locate near to the ground truth corresponding point to ${\bf p}^s_k$, they are served as the \textit{hard} negative examples. The second portion is composed of $N^{-}_k$ samples that are randomly select from the points in ${\cal P}^t$ that are at least $\tau_{hard-}$ from ${\hat{\bf p}}^s_k$. A visual illustration of the sampling process is shown in Fig.~\ref{fig:sampling}. In this manner, $N^{-}_k+N^{hard-}_k$ triplets are sampled for each point in the overlapping region of a shape pair.

\subsection{Implementation details}
The proposed method was implemented in TensorFlow v1.8~\cite{abadi1603tensorflow}. The training samples are permuted and partitioned in minibatches with a size of 512. We use the Adam~\cite{kinga2015method} optimizer with $\beta_1=0.99$ and $\beta_2=0.999$ to train the network. The learning rate is set to 0.0001. The initial weights of the nodes in FC layers follow a normal distribution with a mean of 0 and a standard deviation of 0.1.
\section{Experiments}\label{sec:exp}
In this section, we present the deployed experiments to verify the effectiveness of the proposed method from two perspectives, i.e., feature matching and geometric registration. The following parameter settings are used throughout all the experiments: $\tau_{tri}=1$ and $\tau_{pair}=0.02$ (Eq.~\ref{eq:loss}); $\tau_{anc}=1.5$\textit{pr}, $\tau_{+}=3$\textit{pr} and  $\tau_{hard-}=6$\textit{pr} (Sect.~\ref{subsec:data}); $N_k^-=25$ and $N_k^{hard-}=15$ (Sect.~\ref{subsec:data}). Here, \textit{pr} denotes point cloud resolution, i.e., the average shortest distance among neighboring points computed for a whole dataset.

All the experiments are done on a laptop with a 4-core Intel i7-6700HQ CPU, a 24GB memory, and an NVIDIA GTX1060 graphics card enabled by CUDA v9.0 and cuDNN v7.0.4. The compared methods, feature matching, and geometric registration are implemented based on Point Cloud Library (PCL)~\cite{rusu20113d}.
\subsection{Setup}
\subsubsection{Datasets}
\begin{figure}[htbp]
	\centering
	\includegraphics[width=0.8\linewidth]{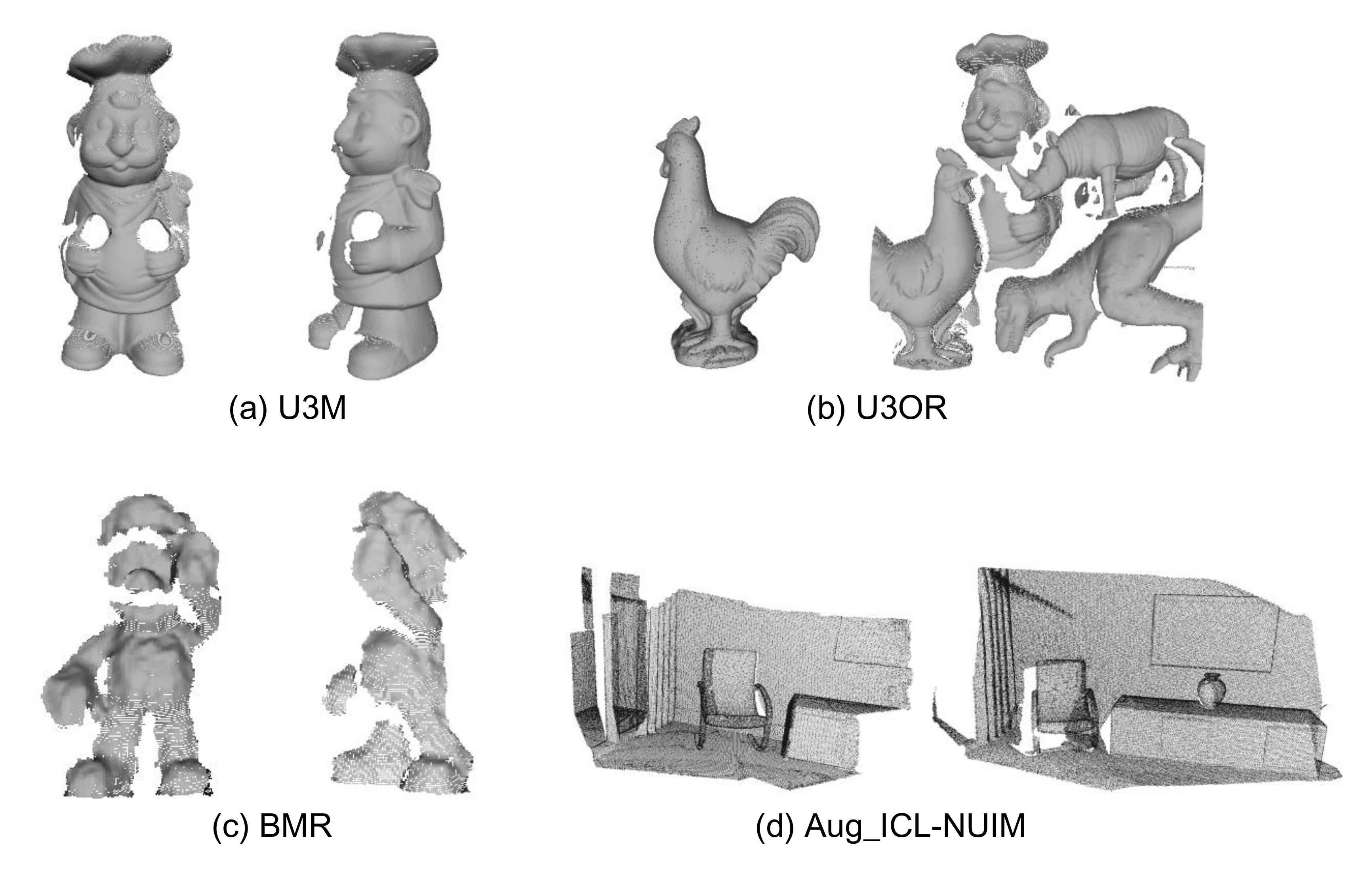}\\
	\caption{Sample views of matching data pairs from four experimental datasets.}
	\label{fig:dataset}
\end{figure}
Four well-known 3D rigid matching datasets with different application scenarios and data modalities are considered for experiments. Some samples from these datasets are shown in Fig.~\ref{fig:dataset}.
\\
\\
\noindent\textbf{U3M~\cite{mian2006novel}}. It is a LiDAR registration dataset that incorporates 22, 16, 16, and 21 partial views of four 3D objects, namely the \textit{Chef}, \textit{Chicken}, \textit{Parasaurolophus}, and \textit{T-Rex}. This dataset is acquired using a Minolta Vivid 910 Scanner. Particularly, we consider any two partial views of the same object with a minimum of 30\% overlaps, resulting in a total of 313 valid matching pairs in this dataset. The ground truth motion between two views is obtained by first manual alignment and then ICP refinement~\cite{besl1992method}.
\\
\\
\noindent\textbf{U3OR~\cite{mian2006three,mian2010repeatability}}. This dataset addresses the 3D object recognition scenario and is very popular for local shape descriptor assessment~\cite{guo2016comprehensive}. It consists of 5 3D models, i.e., \textit{Chef}, \textit{Chicken}, \textit{Parasaurolophus}, \textit{T-Rex}, and \textit{Rhino}, and 50 scenes. The scenes are obtained by randomly placing four or five objects together and scan them from a particular view using a Minolta Vivid 910 Scanner. The challenges provided by U3OR are mainly various degrees of clutter and occlusion. There are 188 valid matching pairs in this dataset. Remarkably, the \textit{Rhino} model is used to create clutter and matching pairs are only found for the other four objects.
\\
\\
\noindent\textbf{BMR~\cite{tombari2010unique}}. BMR is a low-resolution dataset collected using a Microsoft Kinect sensor. It is composed of 15, 16, 20, 13, 16, and 15  partial views of 6 objects, i.e., \textit{Doll}, \textit{Duck}, \textit{Frog}, \textit{Mario}, \textit{PeterRabbit}, and \textit{Squirrel}. Analogously to the U3M dataset, we consider shape pairs with at least 30\% overlaps and eventually get 321 valid matching pairs.
\\
\\
\noindent\textbf{Aug\_ICL-NUIM~\cite{choi2015robust}}. The Augmented
ICL-NUIM (Aug\_ICL-NUIM) dataset was collected via a Microsoft Kinect sensor with four indoor scene sequences including  \textit{Living room 1}, \textit{Living room
2}, \textit{Office 1} and \textit{Office 2}. These 2.5D sequences include 57, 47, 53, and 50 point clouds, respectively. Different from widely-used object datasets, Aug\_ICL-NUIM contains far less geometric details. Fragment pairs in the four sequences that are not temporally adjacent to each other are considered for experiments~\cite{choi2015robust,zhou2016fast}.

In our experiments, the U3M and U3OR datasets are widely employed for the evaluation feature matching performance~\cite{guo2013rotational,yang2017RCS_jrnl} so we use the two datasets for feature matching evaluation; the lager-scale BMR and Aug\_ICL-NUIM datasets are considered for geometric registration evaluation. For U3M and U3OR, matching pairs of \textit{Chef} and \textit{Chicken}, \textit{Parasaurolophus}, and \textit{T-Rex} are employed for training, validation, and testing, respectively. For the BMR dataset, we use the matching pairs of \textit{Doll}, \textit{Duck}, and \textit{Frog} for training and the remaining for testing. For the Aug\_ICL-NUIM dataset, we follow~\cite{khoury2017learning} and train our network on the SceneNN~\footnote{https://marckhoury.github.io/CGF/} dataset. Remarkably, there are benchmark results of several deep learned descriptors for the Aug\_ICL-NUIM dataset, making it possible to fairly compare our fused descriptors with deep learned descriptors .

\subsubsection{Criteria}
\begin{table}[t]\small
	\renewcommand{\arraystretch}{1}
		\caption{Network settings for fusing different features.}
	\centering
	\begin{tabular}{lccc}
		\hline
		& $N_{intra}$ & $N_{inter}$ & \# epochs\\
		\hline
		LFSH~\cite{yang2016fast}&48&256&3\\
		RCS~\cite{yang2017RCS_jrnl}&48&512&3\\
		SI~\cite{johnson1999using}+SHOT~\cite{tombari2010unique}&512&512&5\\
		SI~\cite{johnson1999using}+SHOT~\cite{tombari2010unique}+RCS~\cite{yang2017RCS_jrnl}&512&512&5\\
		\hline
	\end{tabular} 
	\label{tab:para}
\end{table}
We use the Recall versus Precision curve (RPC) as suggested by many previous works~\cite{tombari2010unique,guo2013rotational} to examine the feature matching performance of a local shape descriptor. It is calculated as follows: given a source shape, a target shape, and the ground truth transformation, we randomly sample 1000 keypoints for the source shape and locate their corresponding points in the target shape using the ground truth transformation. Local shape features are extracted for the keypoints and a target feature is matched against all source features to find the closest feature. If the ratio of the smallest distance to the second feature distance is smaller than a threshold, the target feature and the closest source feature is considered as a match. A match is further considered as correct if it conforms to the ground truth transformation. Precision refers to the ratio of the number of feature-identified correct matches to the number of matches. Recall refers to the ratio of the number of feature-identified correct matches to the number of total correct matches. We will also present the area under curve (AUC) values for each RPC to measure the overall feature matching performance.
\begin{figure}[htbp]
	\begin{minipage}{0.495\linewidth}
		\centering
		\subfigure[LFSH on U3M]{
			\includegraphics[width=1.0\linewidth]{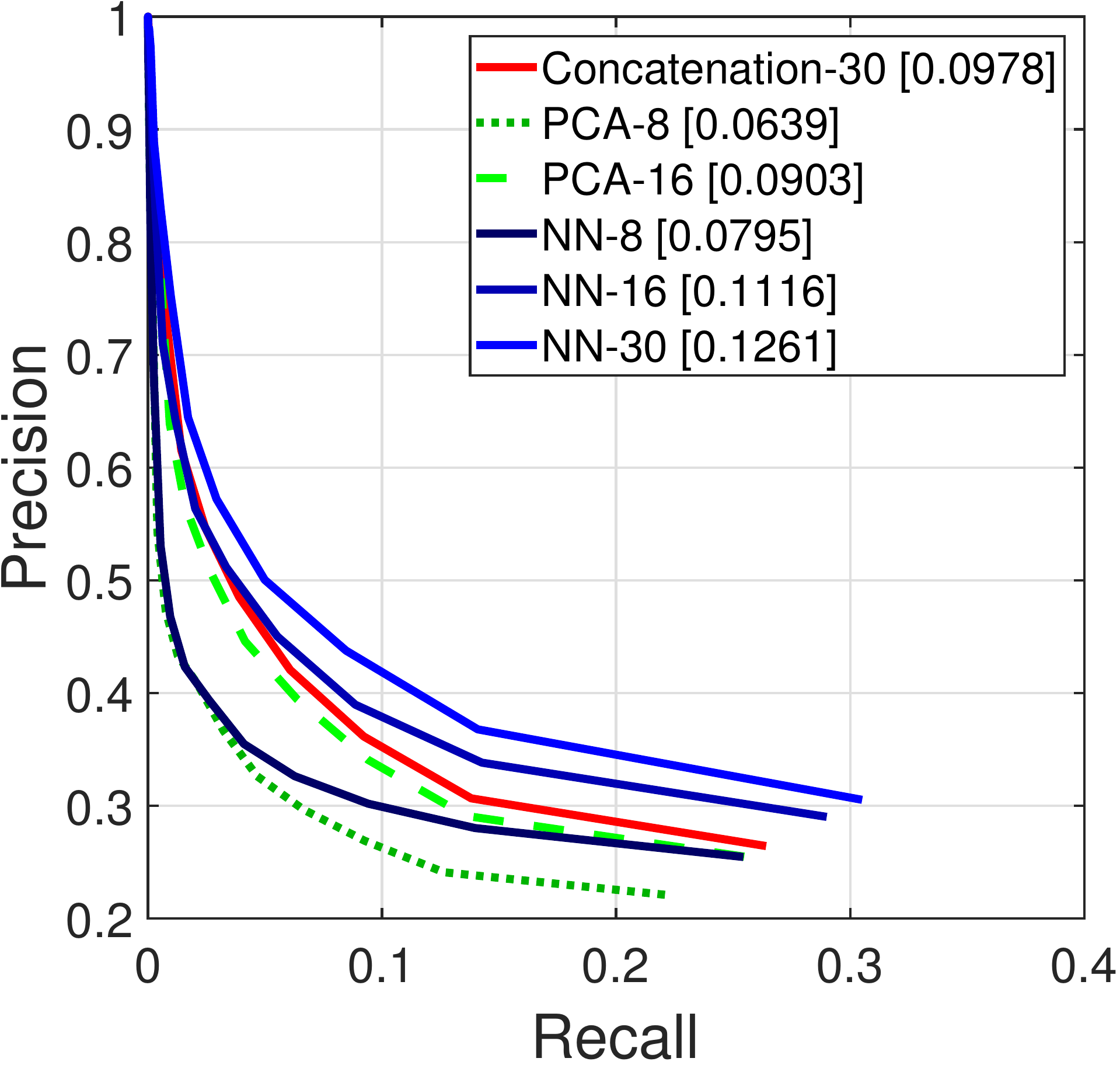}}
	\end{minipage}
	\hfill
	\begin{minipage}{0.495\linewidth}
		\centering
		\subfigure[LFSH on U3OR]{
			\includegraphics[width=1.0\linewidth]{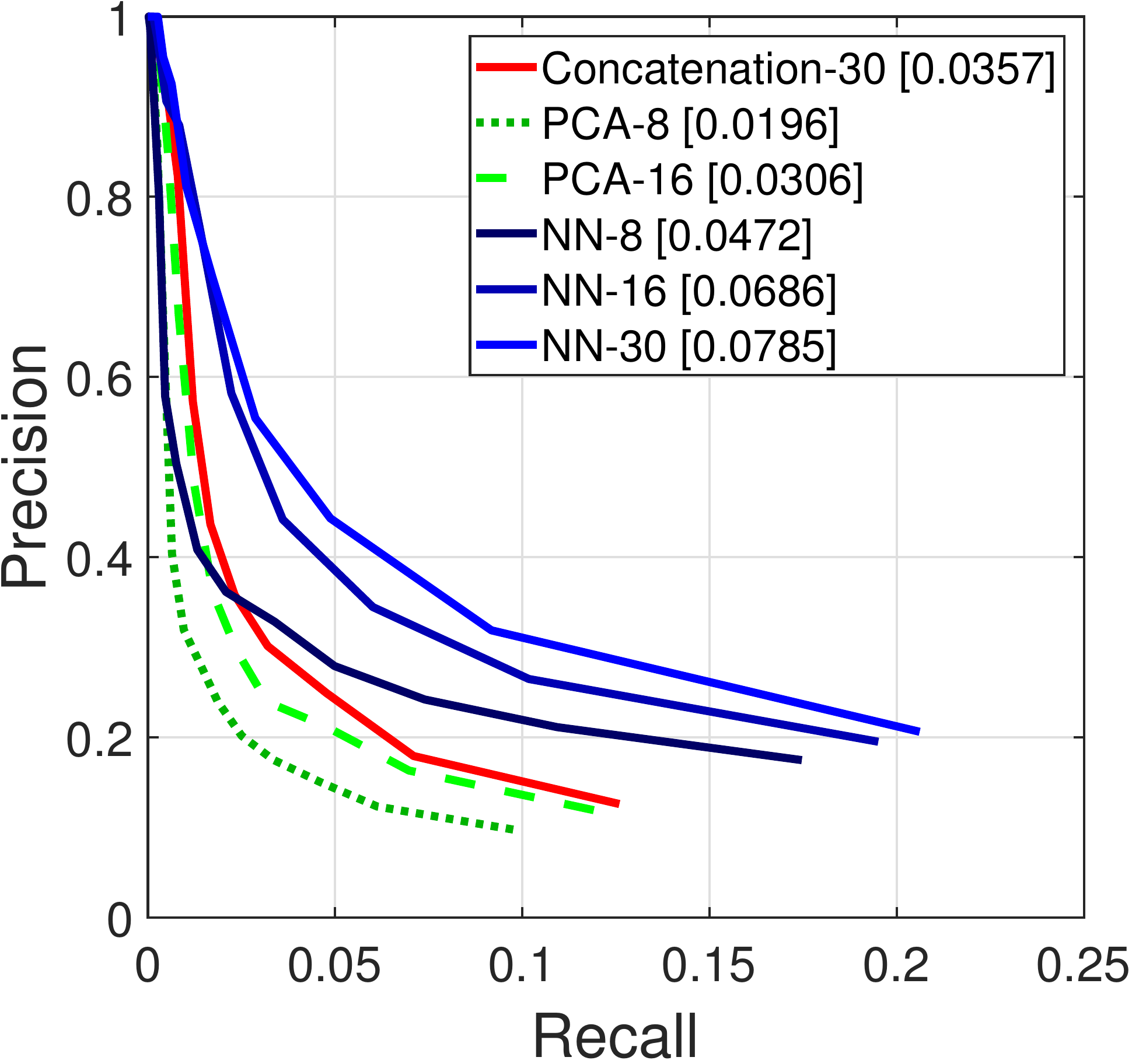}}
	\end{minipage}
	\hfill
	\begin{minipage}{0.495\linewidth}
		\centering
		\subfigure[RCS on U3M]{
			\includegraphics[width=1.0\linewidth]{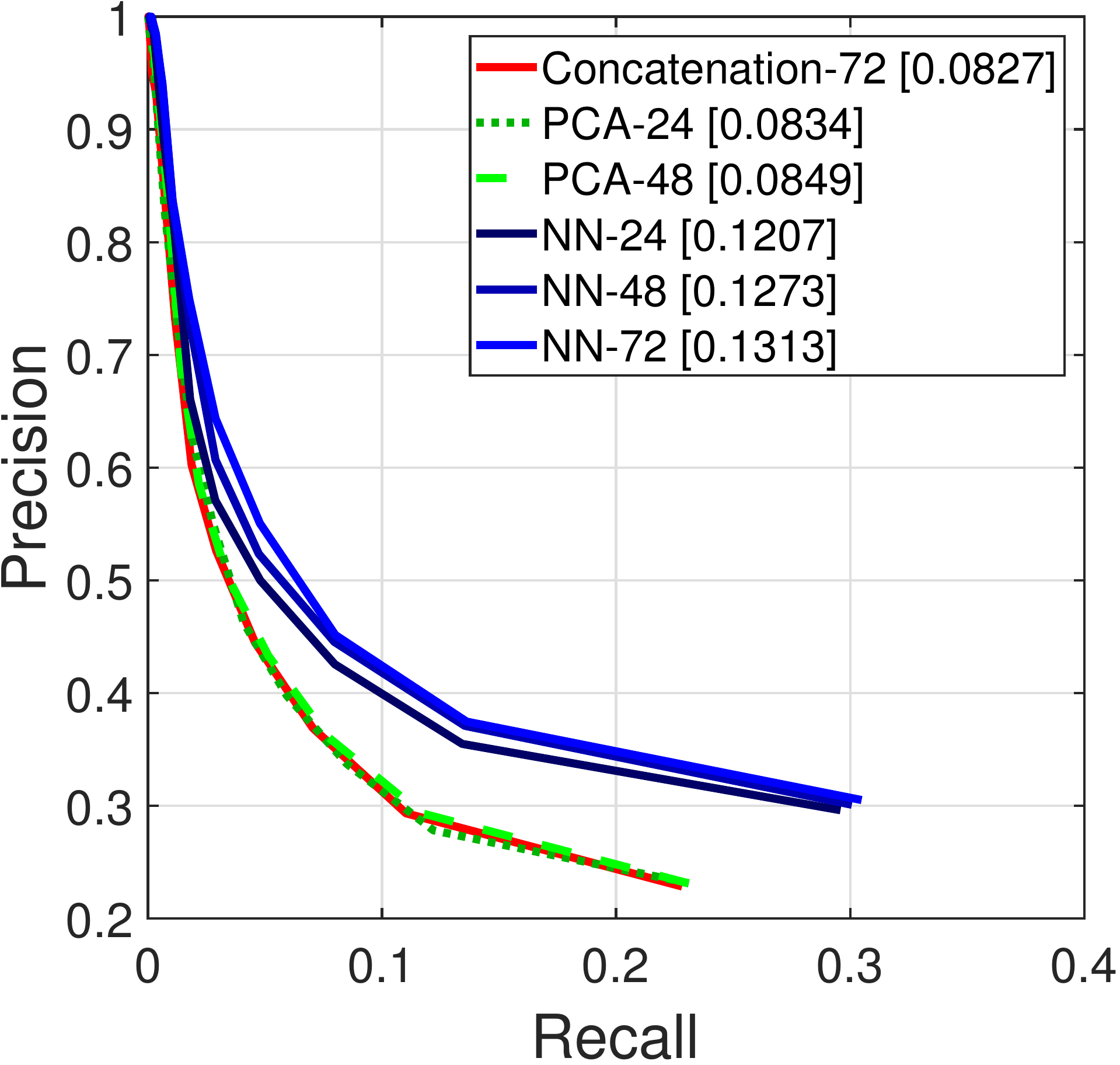}}
	\end{minipage}
	\hfill
	\begin{minipage}{0.495\linewidth}
		\centering
		\subfigure[RCS on U3OR]{
			\includegraphics[width=1.0\linewidth]{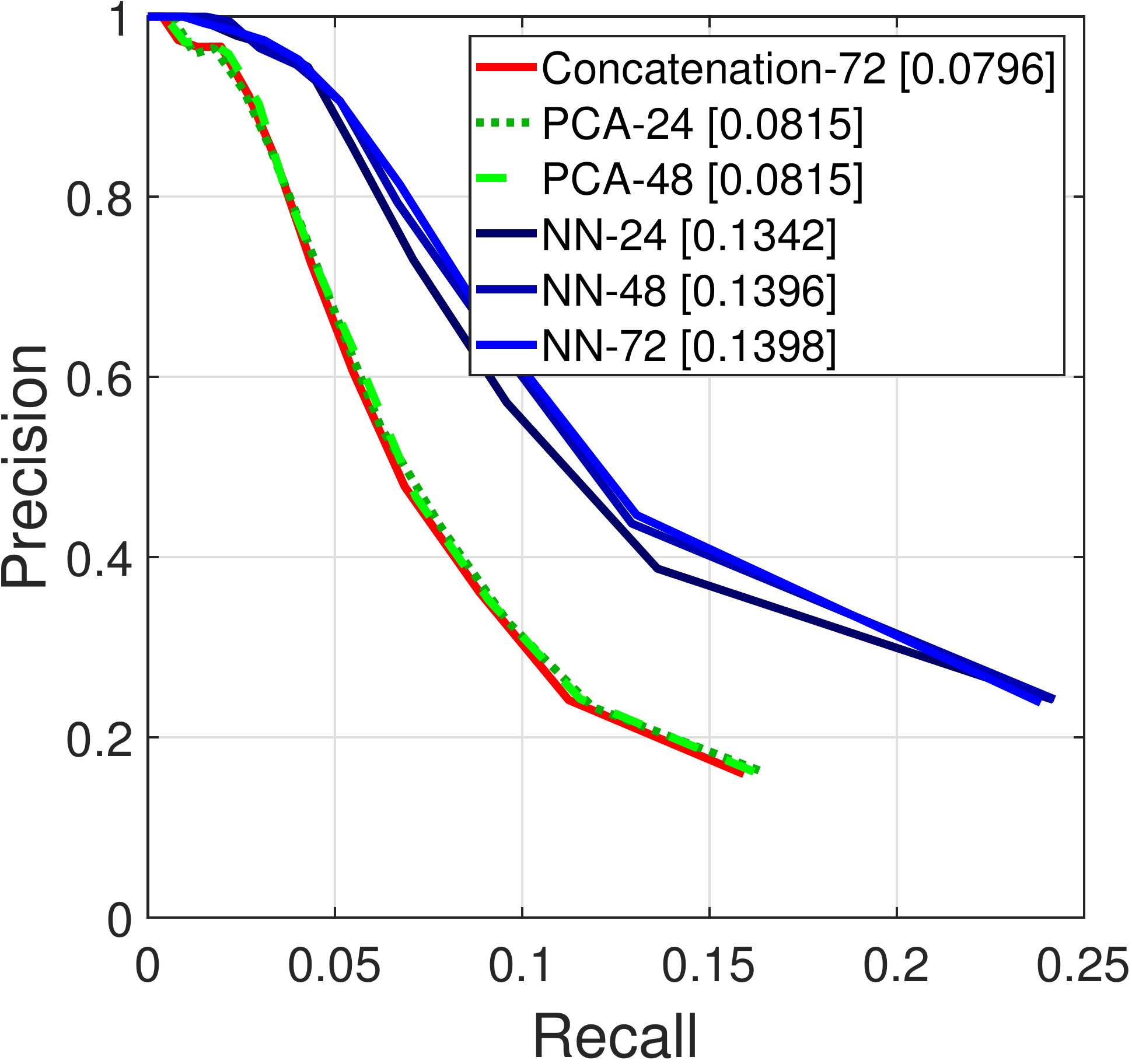}}
	\end{minipage}
	\hfill
	\caption{Feature matching performance of local geometric descriptors generated by different low-level feature fusion techniques on the test sets of the U3M and U3OR datasets. Here and hereinafter, numbers in square brackets are AUC values and numbers attached behind each fusion method indicate the dimensionality of fused features.}
	\label{fig:low-level_result}
\end{figure}

Regarding geometric registration performance assessment, we employ the $\alpha$-recall suggested in~\cite{zhou2016fast} for the BMR dataset. It is defined as the fraction of success registrations whose RMSE values are smaller than $\alpha$. RMSE is computed on the distances between ground truth correspondences after registration~\cite{yang2016fast,choi2015robust}, which is defined as:
\begin{equation}\label{eq:rmse}
\begin{aligned}
{\rm {RMSE}}({\cal P}^s, {\cal P}^t)=   \frac{1}{{|C^*|}}\sum\limits_{({{\bf p}^*_s},{{\bf p}^*_t}) \in C^*} {||{{\bf{R}}_{GT}}{{\bf p}^*_s} +{\bf t}_{GT} - {{\bf p}^*_t}|{|^2}}  ,
\end{aligned}
\end{equation}
where $C^*$ is the ground-truth correspondence set between ${\cal P}^s$ and ${\cal P}^t$; ${\bf R}_{ij} \in SO(3)$ and ${\bf t}_{ij} \in \mathbb{R}^3$ are ground-truth rotation and translation, respectively. Regarding the Aug\_ICL-NUIM dataset, we employ the benchmarking metrics, i.e., \textit{Precision} and  \textit{Recall} of successfully aligned data pairs (we additionally consider the \textit{F-score}, i.e., $2\frac{Precision\cdot Recall}{Precision+Recall}$,  as an aggregated metric). Specifically, a registration  is considered as correct if its RMSE (Eq.~\ref{eq:rmse}) is smaller than 0.2~\cite{choi2015robust}. 
\subsubsection{Features for fusion}
For low-level local geometric feature fusion, we consider the LFSH~\cite{yang2016fast} (30 dim.) and RCS~\cite{yang2017RCS_jrnl} (72 dim.). LFSH is the concatenation of three sub-histograms that encode signed distance, normal deviation, and density attributes, respectively. RCS is the concatenation of six signatures that represents the contour information from different views.

Regarding high-level feature fusion, two combinations are considered: spin image (SI)~\cite{johnson1999using} (153 dim.)+SHOT~\cite{tombari2010unique} (352 dim.), and SHOT~\cite{tombari2010unique} (352 dim.)+RoPS~\cite{guo2013rotational} (135 dim.)+RCS~\cite{yang2017RCS_jrnl} (72 dim.). A common trait of our considered combination is that features within the combination encode local geometry from different perspectives, which are supposed to be complementary to each other.

We set the support radius of all descriptors to 15\textit{pr} on the U3M, U3OR, and BMR datasets. For the Aug\_ICL-NUIM dataset with large-scale indoor data, 60\textit{pr} is used to include more structure details. The network parameters used to fuse different features are reported in Table~\ref{tab:para}.
\subsubsection{Compared methods}
Existing methods for low-level geometric feature fusion, i.e., concatenation~\cite{tombari2010unique,yang2016fast} and PCA~\cite{johnson1999using,guo2015novel}, are taken into comparison. Specifically, PCA is trained over a set of descriptors computed for each point of the \textit{Chef} and \textit{Chicken} models. For high-level geometric feature fusion, we compare our method with the min pooling approach~\cite{buch2016local}. We also investigate the effectiveness of applying concatenation and PCA for descriptor-level fusion.

Because the proposed method is a fusion approach, thus we mainly focus on the comparison with existing fusion methods for local geometric features, but we will still include comparisons with some traditional and deep learned descriptors.
\subsection{Low-level feature fusion performance}
\begin{figure}[htbp]
	\begin{minipage}{0.495\linewidth}
		\centering
		\subfigure[SI+SHOT on U3M]{
			\includegraphics[width=1.0\linewidth]{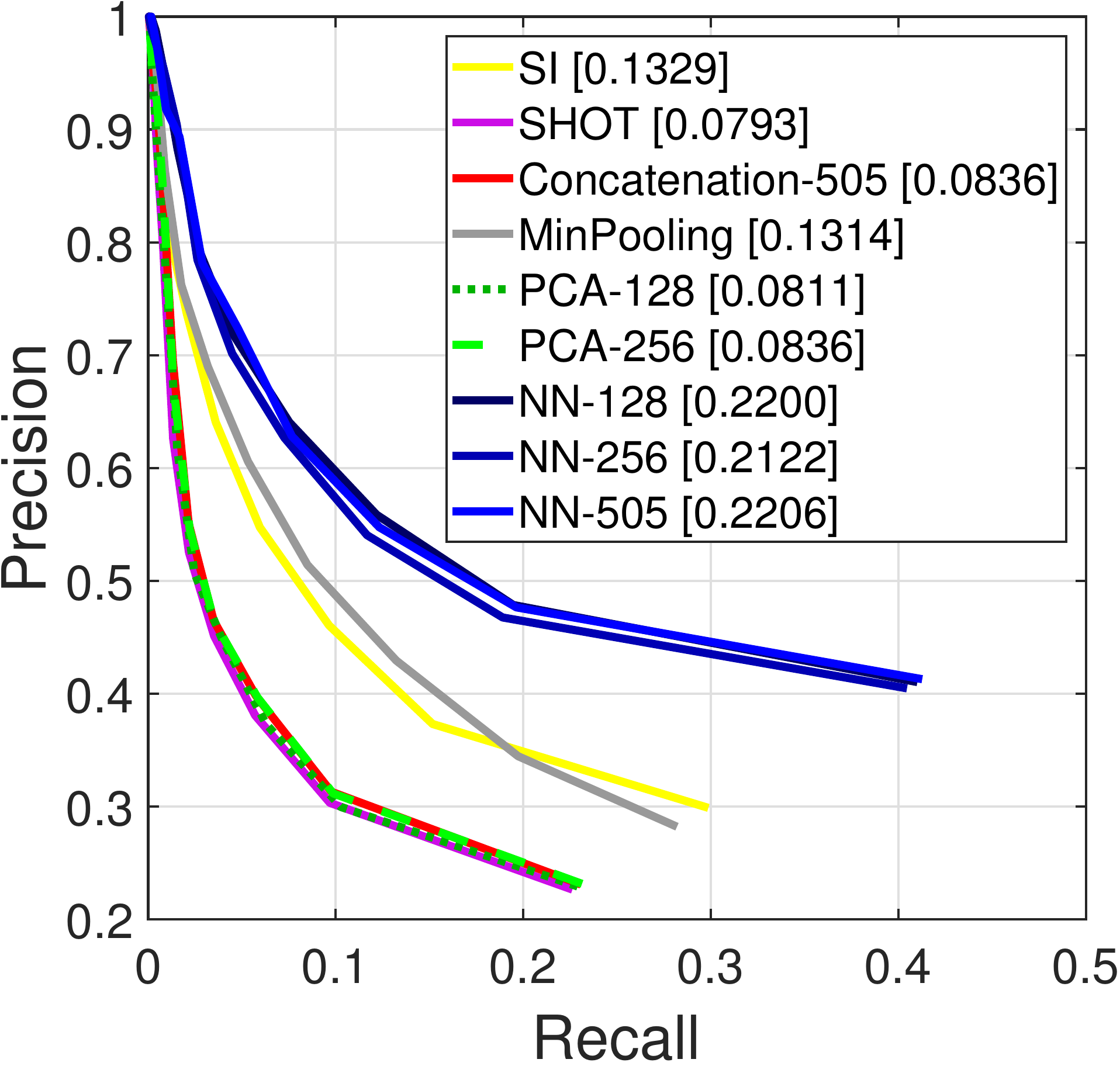}}
	\end{minipage}
	\hfill
	\begin{minipage}{0.495\linewidth}
		\centering
		\subfigure[SI+SHOT on U3OR]{
			\includegraphics[width=1.0\linewidth]{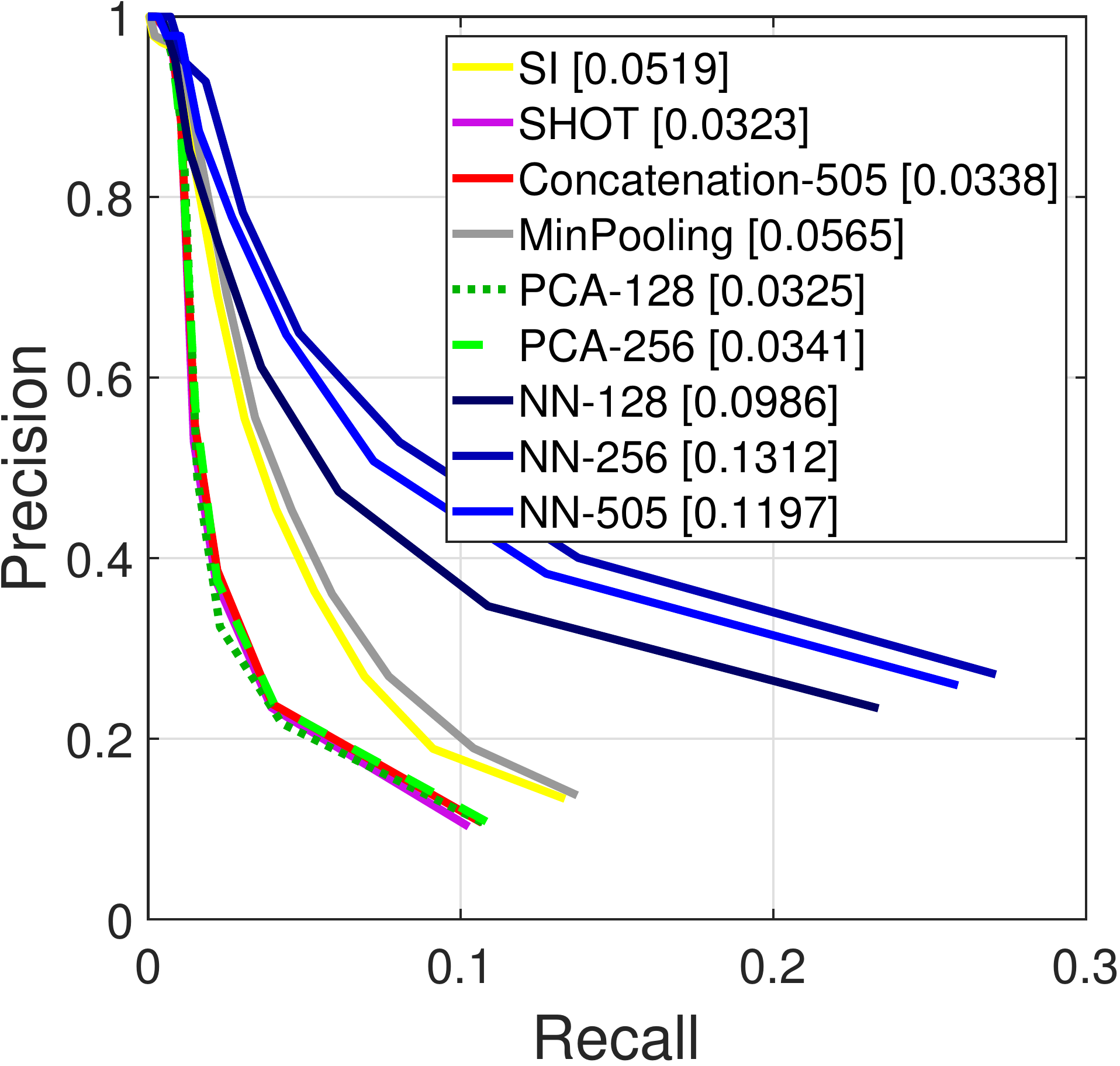}}
	\end{minipage}
	\hfill
	\begin{minipage}{0.495\linewidth}
		\centering
		\subfigure[SHOT+RoPS+RCS on U3M]{
			\includegraphics[width=1.0\linewidth]{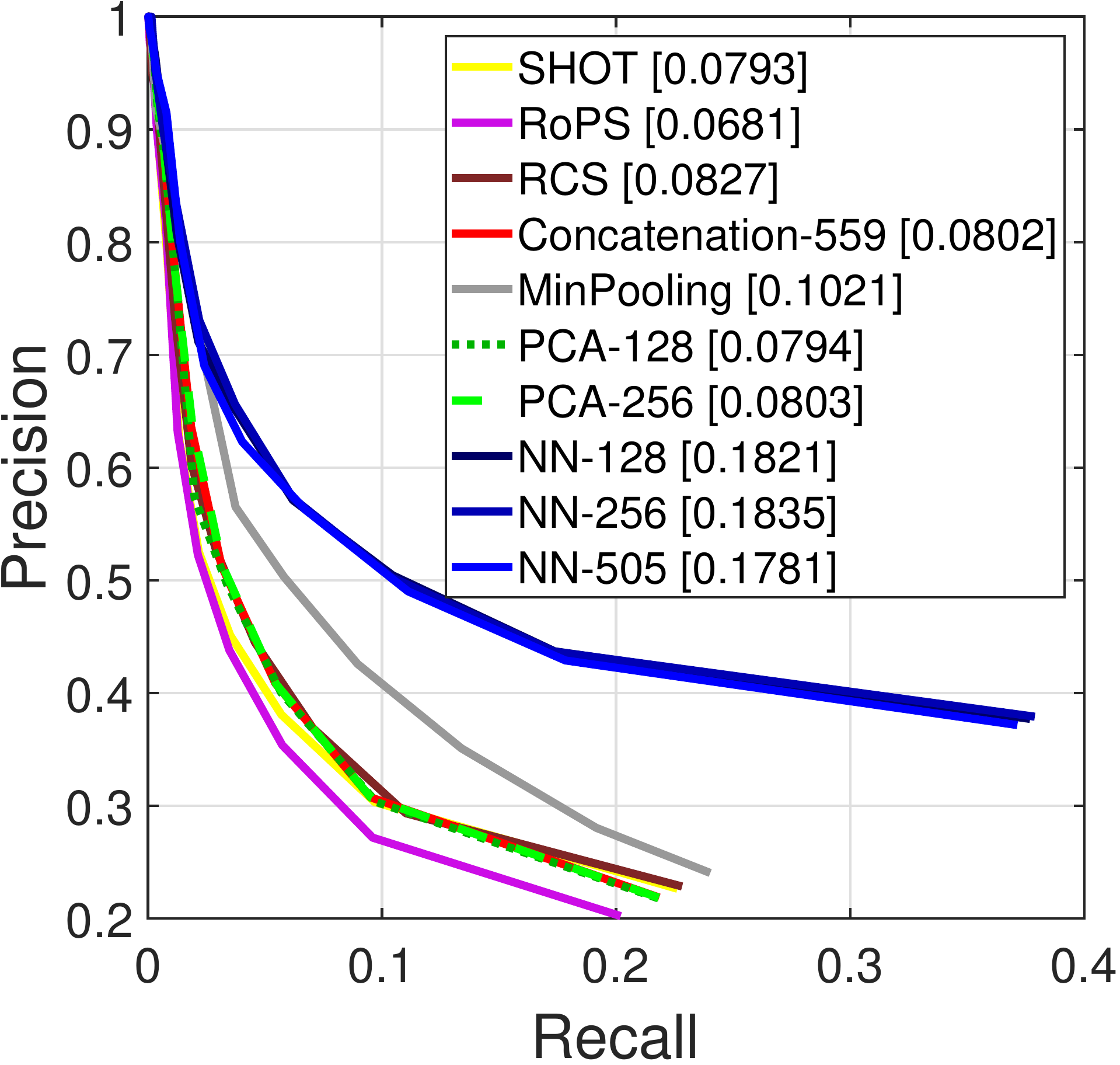}}
	\end{minipage}
	\hfill
	\begin{minipage}{0.495\linewidth}
		\centering
		\subfigure[SHOT+RoPS+RCS on U3OR]{
			\includegraphics[width=1.0\linewidth]{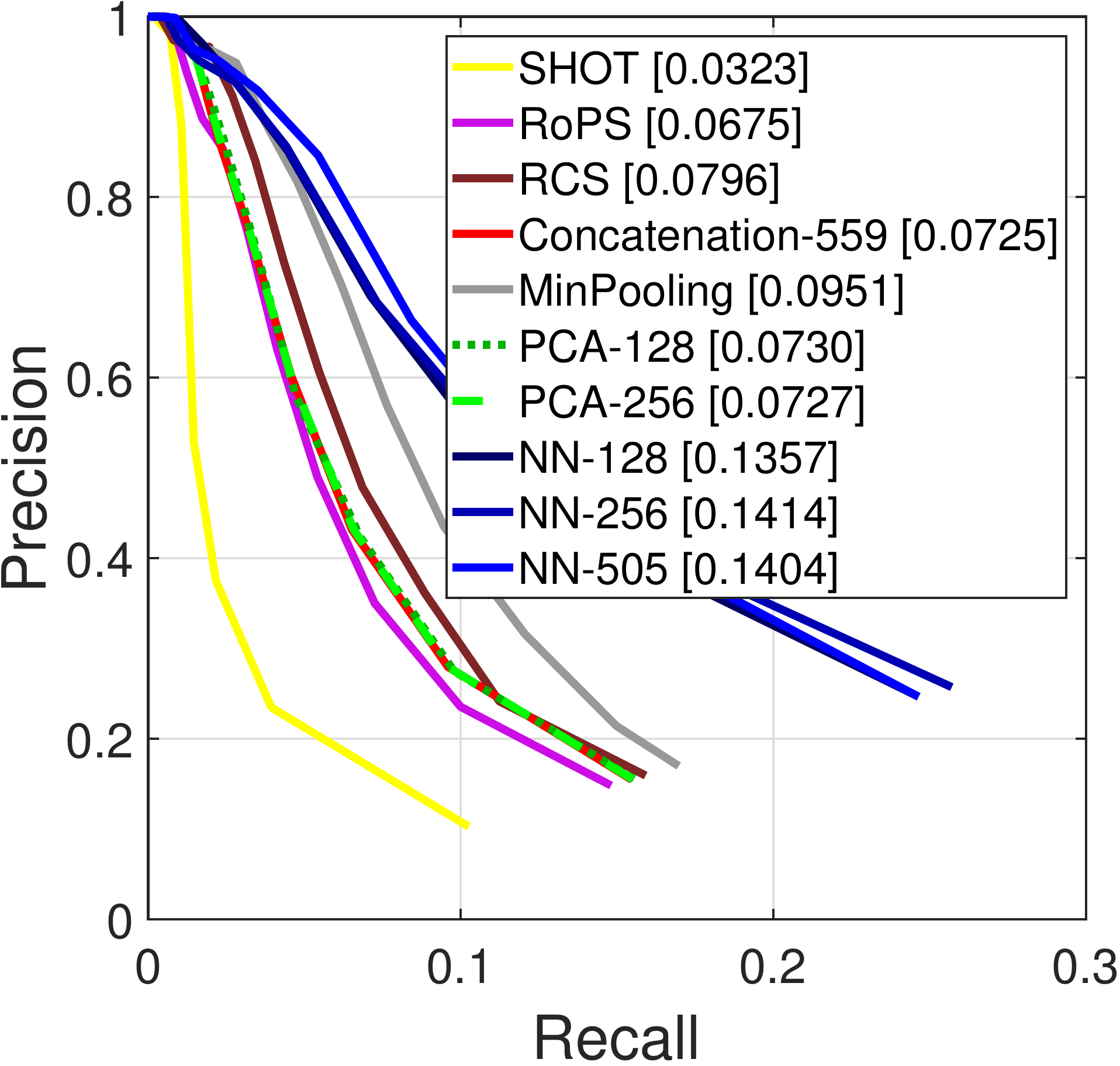}}
	\end{minipage}
	\hfill
	\caption{Feature matching performance of different local geometric descriptors and their combinations by several high-level feature fusion techniques on the test sets of the U3M and U3OR datasets.}
	\label{fig:high-level_result}
\end{figure}
We present the feature matching performance of local geometric descriptors created by our fusion method (dubbed as NN) as well as the compared approaches on the test sets of the U3M and U3OR datasets in Fig.~\ref{fig:low-level_result}. Different feature lengths are specifically examined for PCA and NN to see whether a good balance could be achieved between distinctiveness and compactness. 

For fusion results on LFSH in Fig.~\ref{fig:low-level_result}(a)-(b), one can see that the proposed method manages to achieve better performance than the original LFSH feature (i.e., generated via histogram concatenation) with 16 and 30 dimensions. The gap is especially significant  on the U3OR dataset. Notably, our NN-fused LFSH with mere 8 dimensions outperforms the original version on the U3OR dataset. Although LFSH is known to be fast and compact~\cite{yang2016fast}, NN can further improve its compactness while making it more discriminative (the NN fusion process was performed in real time). PCA, though making the LFSH feature more compact, results in a slight performance deterioration. For the RCS feature, consistent results can be witnessed in Fig.~\ref{fig:low-level_result}(a)-(b). Specifically, our NN method surpasses all methods in terms of both compactness and distinctiveness by a clear margin on the U3M and U3OR datasets. These results demonstrate that integrating features non-linearly by matching-guided learning  is able to fully exploit the advantageous components to feature matching existed in the feature set and aggregate them in a compact manner.
\subsection{High-level feature fusion performance}
\begin{table}[htbp]\small
	\renewcommand{\arraystretch}{1}
	\centering
	\caption{AUC values of learned feature fusions with different loss functions (i.e., contrastive loss~\cite{zeng20173dmatch}, triplet loss~\cite{khoury2017learning}, and our loss), network architectures, and inputs on the validation sets of the U3M and U3OR datasets. ``w/o intra.'' indicates removing the intra-feature fusion block and using the concatenated feature vector as the input to fully connected layers, as done in~\cite{khoury2017learning}. $N^\star$ represents the best performance when using any $N$ features for fusion from the prepared feature set. The learned features from (a) to (b) have 16, 48, 256, and 256 dimensions, respectively. The best results are in bold fonts.}
	\begin{tabular}{@{\extracolsep{\fill}}lccc}
		\hline
		& U3M & U3OR& Avg.\\
		\hline
		\textit{(a) LFSH} &&&\\
		w/o intra.& 0.0865& 0.0369& 0.0617 \\
		Contrastive loss& 0.0365& 0.0245 & 0.0305\\
		Triplet loss&\bf 0.1044 & 0.0586 & \bf 0.0815 \\
		Ours& 0.1023& \bf 0.0596 & 0.0810\\
		\hline
		\textit{(b) RCS} &&&\\
		w/o intra.& 0.1126 & 0.1255 & 0.1191\\
		Contrastive loss& 0.0521 & 0.0582 & 0.0552\\
		Triplet loss& 0.1256 & \bf 0.1498 & 0.1377\\
		Ours& \bf 0.1269 & 0.1486 & \bf 0.1378\\
		\hline
		\textit{(c) SI+SHOT} &&&\\
		1$^\star$& 0.1879  & 0.1158 & 0.1519\\
		w/o intra.& 0.2105 & 0.1408 & 0.1757\\
		Contrastive loss& 0.0868 & 0.0329 & 0.0599\\
		Triplet loss& 0.2361 & 0.1207 & 0.1784\\
		Ours& \bf 0.2452 & 0.\bf 1465 & \bf 0.1959 \\
		\hline
		\textit{(d) SHOT+RoPS+RCS} &&&\\
		1$^\star$& 0.1056 & 0.0774 & 0.0915\\
		2$^\star$& 0.1799 & 0.1142 & 0.1471\\
		w/o intra.& 0.1910  & 0.1211 & 0.1561\\
		Contrastive loss& 0.0569 & 0.0398 &0.0484 \\
		Triplet loss& 0.1785 & 0.1129 & 0.1457\\
		Ours& \bf 0.1957 & \bf 0.1325 & \bf 0.1641\\
		\hline
	\end{tabular} 
	
	\label{tab:analysis}
\end{table}
Fig.~\ref{fig:high-level_result} shows the feature matching performance of local geometric descriptors and their fused results by several high-level feature fusion methods. For the fusion of SI and SHOT (Fig.~\ref{fig:high-level_result}(a)-(b)), NN achieves the best performance on both datasets under all tested dimensions, outperforming the second best, i.e., min pooling, by a significant margin. Concatenation and PCA behave poorly. The reason can be reflected by the phenomenon that SI exhibits comparative performance with min pooling, indicating that in most patch matching cases SI produces higher similarity score than SHOT. In this sense, attaching SHOT to SI will cause many noisy bins in the concatenated feature (we will show that SHOT actually has a postive impact on SI when fused via NN in Sect.~\ref{sec:analysis}).

Similar conclusion can be drawn from the fusion results of SHOT, RoPS, and RCS as presented in Fig.~\ref{fig:high-level_result}(c)-(d), i.e., NN outperforms all others. Since min pooling returns better performance than using any single feature, we can infer that these descriptors provide  complementary  information to each other. Compared with min pooling, NN  mines such information more effectively. In addition, more time is dedicated to feature matching by min pooling because the feature similarity score needs to be calculated $N$ times ($N$ being the number of descriptors).
\subsection{Method analysis}\label{sec:analysis}

For the purpose of  verifying the effectiveness of our improved triplet loss and the rationality of our network design, we conduct a set of experiments on the validation sets of U3M and U3OR datasets with different loss functions, network architectures, and inputs. The results are reported in Table~\ref{tab:analysis}.

Three main observations can be made from the results. First, the inter-feature fusion block that performs information exchange within a feature is demonstrated to be beneficial for multi-feature fusion as it brings performance gain in all tested cases. Second, recall that directly concatenating all features often lead to inferior performance than using a particular feature as shown in Fig.~\ref{fig:high-level_result}. Yet, under the propoposed NN fusion framework, the best performance is achieved when taking all features as the input. For instance, SI is more distinctive than the concatenated SI+SHOT feature, while NN reveals that SHOT does provide complementary information to SI since a clear performance gain is witnessed when using SI+SHOT for learning. This is somewhat not surprising because SI and SHOT describe the local surface from different perspectives, i.e. density and normal deviations, and intuitively we should expect implicit complementary information within SI+SHOT. Based on this observation, we can conclude that NN is capable of fully mining the complementary information within multiple features. Third, the improved triplet loss, when keeping other settings identical, generally yields better results than the original triplet loss~\cite{khoury2017learning}, showing the effectiveness of our improvement. We also try the  contrastive loss~\cite{zeng20173dmatch} but it turns out to be far inferior to other losses. Similar results have also been reported in~\cite{khoury2017learning} when learning representations from local geometric features.
\subsection{3D rigid geometric registration}
\begin{figure}[htbp]
	\begin{minipage}{0.495\linewidth}
		\centering
		\subfigure[RCS, 200 iters.]{
			\includegraphics[width=1.0\linewidth]{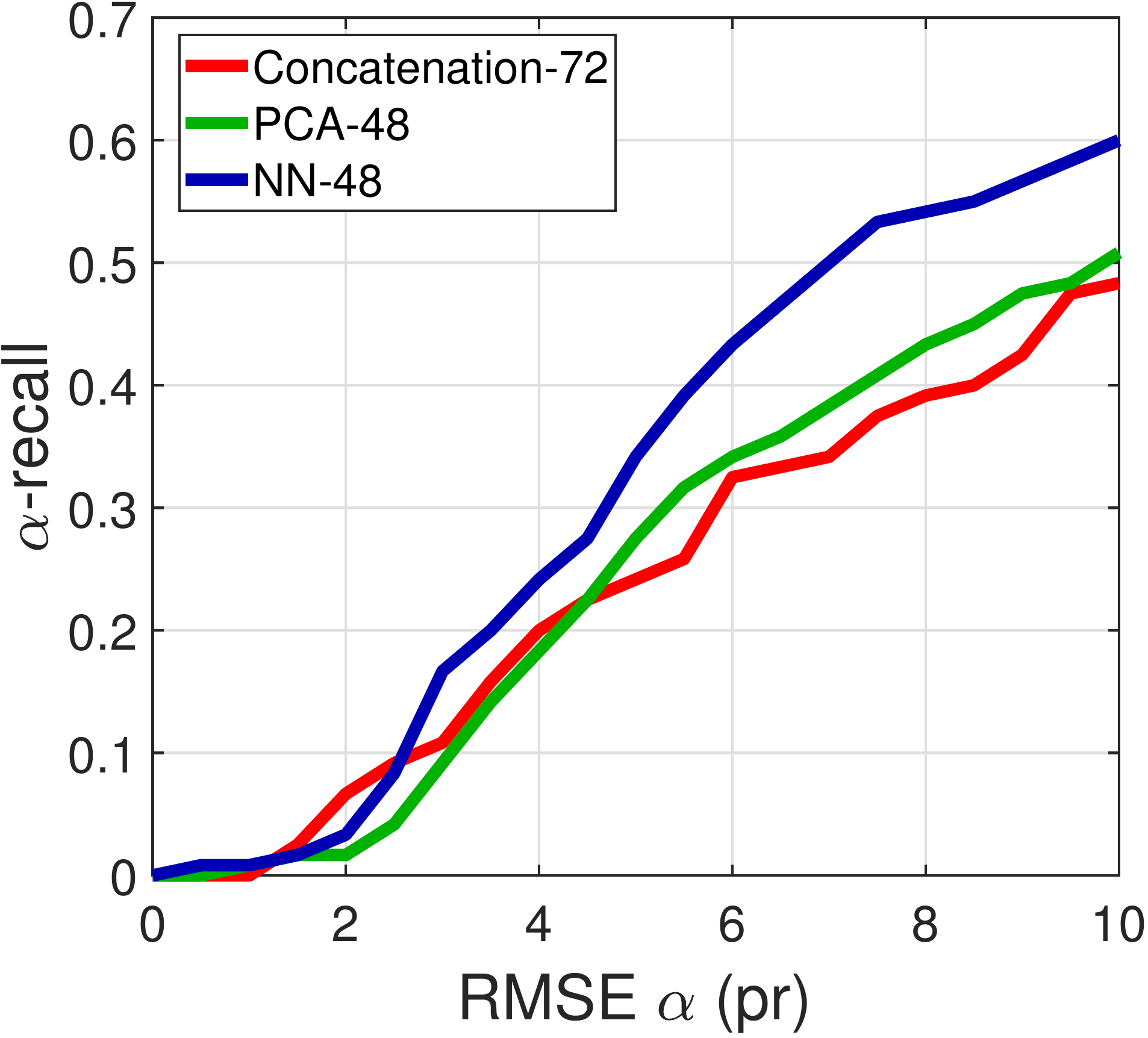}}
	\end{minipage}
	\hfill
	\begin{minipage}{0.495\linewidth}
		\centering
		\subfigure[RCS, 1000 iters.]{
			\includegraphics[width=1.0\linewidth]{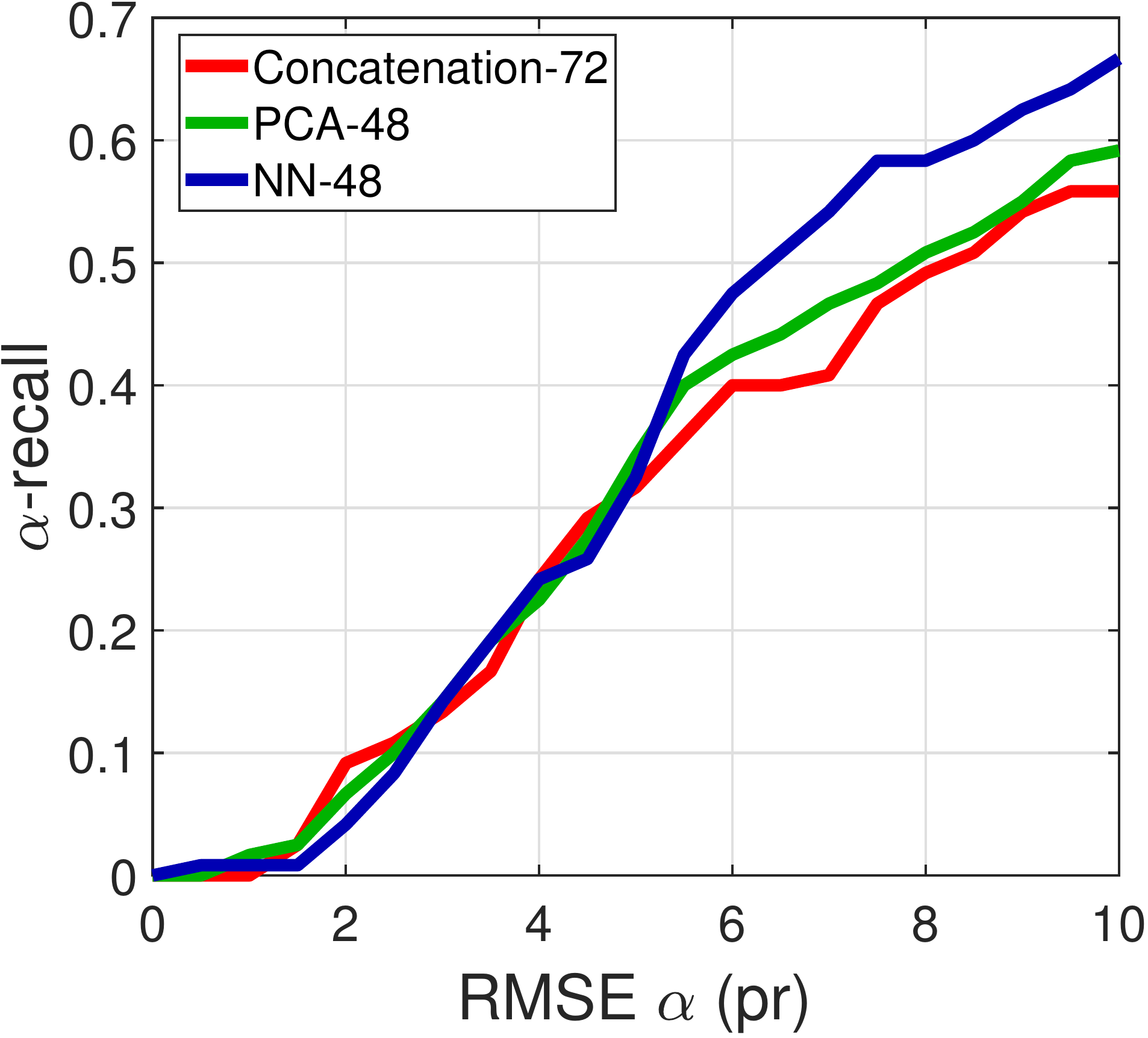}}
	\end{minipage}
	\hfill
	\begin{minipage}{0.495\linewidth}
		\centering
		\subfigure[SHOT+RoPS+RCS, 200 iters.]{
			\includegraphics[width=1.0\linewidth]{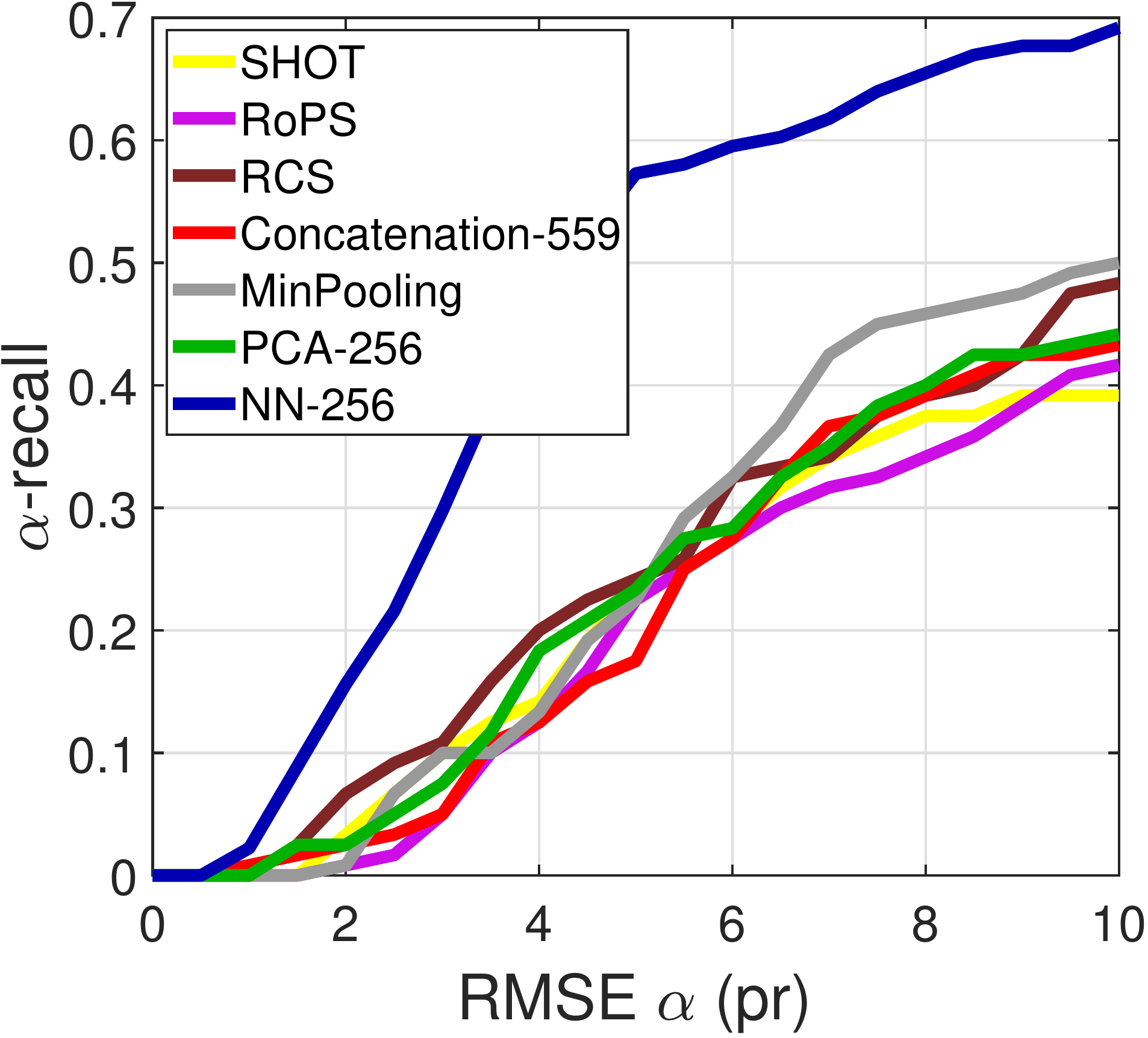}}
	\end{minipage}
	\hfill
	\begin{minipage}{0.495\linewidth}
		\centering
		\subfigure[SHOT+RoPS+RCS, 1000 iters.]{
			\includegraphics[width=1.0\linewidth]{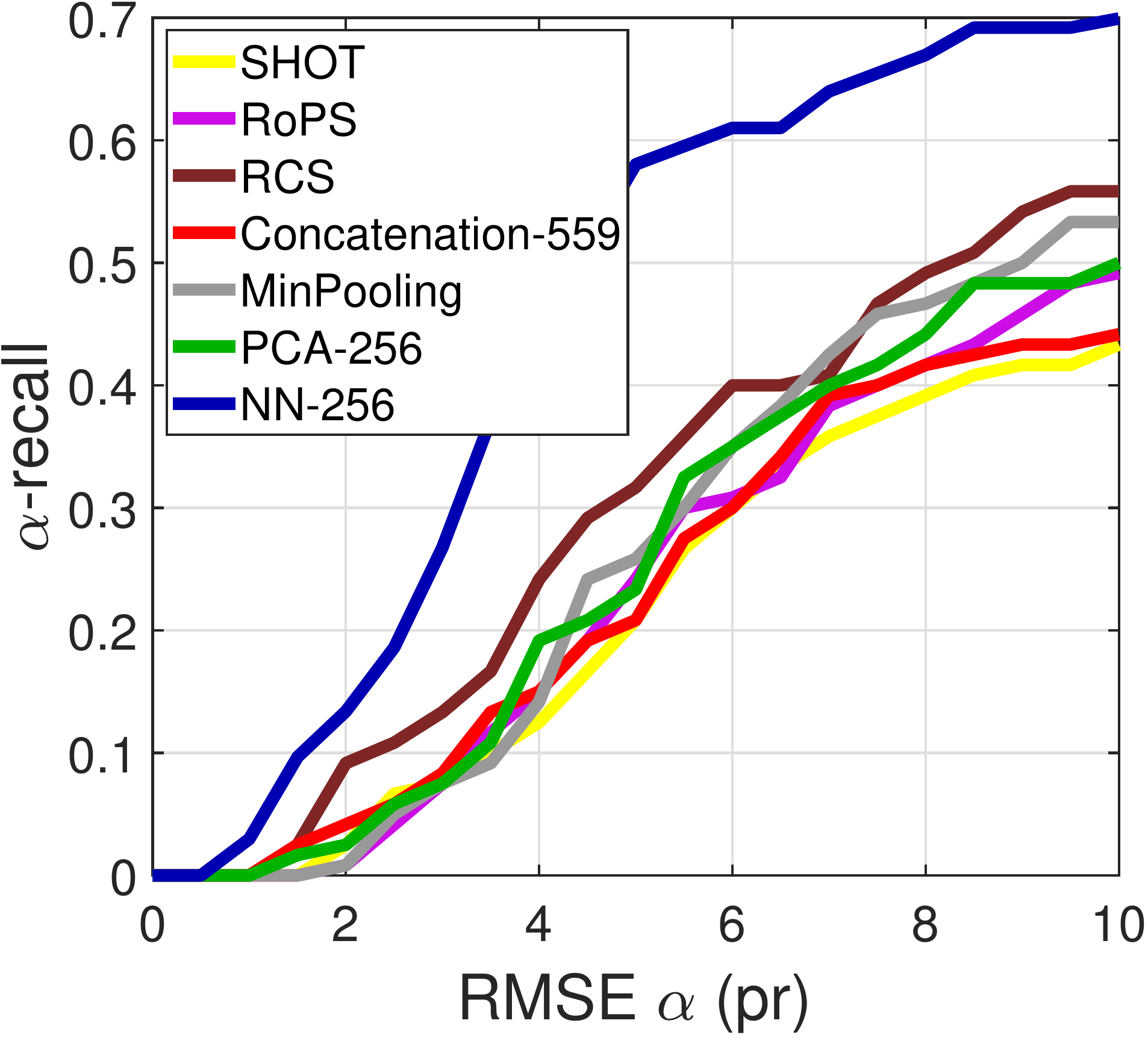}}
	\end{minipage}
	\hfill
	\caption{Quantitative geometric registration results measured by $\alpha$-recall on the test set of the BMR dataset. The registration is performed by a standard local feature matching-based registration pipeline~\cite{holz2015registration} without refinement. We consider 200 and 1000 RANSAC iterations for each tested fusion case.}
	\label{fig:reg_result}
\end{figure}

For task-level evaluation, i.e., 3D rigid registration, we consider RCS and SHOT+RoPS+RCS that respectively belong to low-level and high-level feature fusion scenarios to generate local geometric features. To perform registration, we follow the standard pipeline in PCL~\cite{holz2015registration}, including keypoint detection,  feature description, feature matching, transformation estimation, and refinement. We use uniform sampling provided by PCL to detect keypoints, the fused features by tested methods for feature description, KNN~\cite{muja2014scalable} for feature matching, and RANSAC~\cite{fischler1981random} for transformation estimation. No refinement such as iterative closest points (ICP)~\cite{besl1992method} is performed in this experiment.

\begin{figure}[t]
	\centering
	\includegraphics[width=1.0\linewidth]{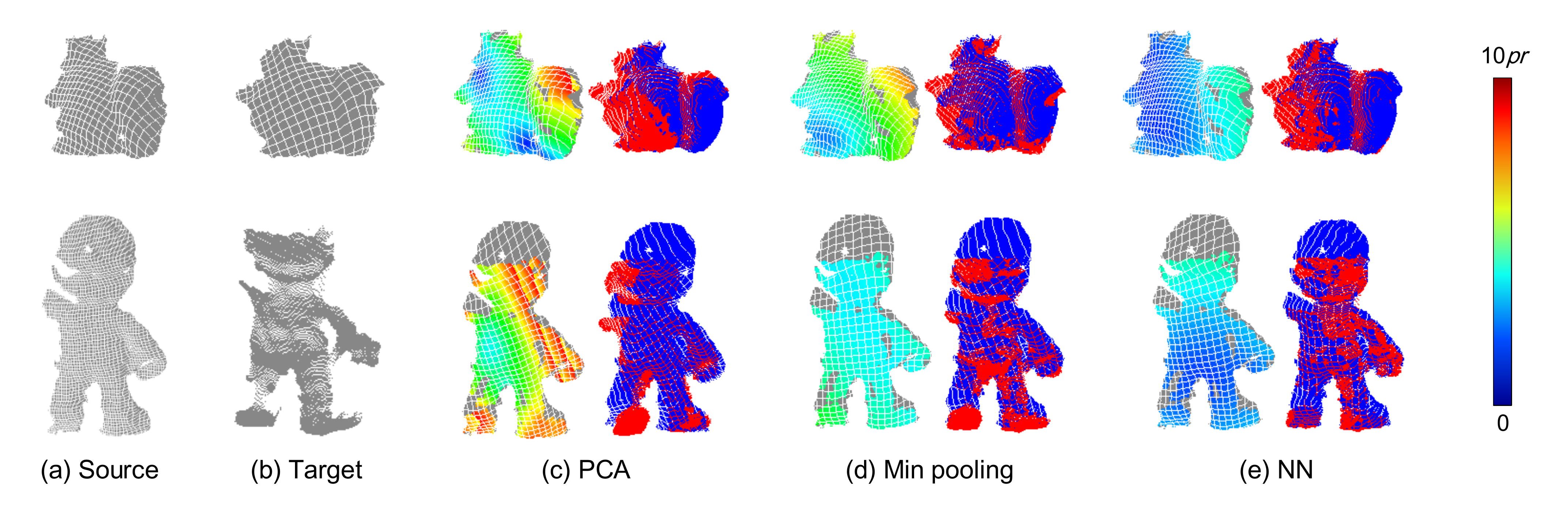}\\
	\caption{Visualization of two exemplar registration results from the BMR dataset using different local geometric descriptors generated by performing PCA, min pooling~\cite{buch2016local}, and NN fusion on SHOT+RoPS+RCS. From left to right: source data, target data, point-wise registration errors and final registration results of PCA, min pooling, and our NN method. From (c) to (d), the source and target point clouds are shown in red and green, respectively; point-wise errors are shown in the source point cloud that are coded by colors presented in the right bar.}
	\label{fig:reg_view}
\end{figure}
\paragraph{\textbf{1) Results on the BMR dataset}}
We consider different RANSAC iterations since this parameter often varies with the quality of feature correspondences and will result in various registration speeds.  The $\alpha$-recall results are presented in Fig.~\ref{fig:reg_result}.

For RCS fusion, NN surpasses other methods with all tested RANSAC iterations. The margin is more obvious with 200 iterations, showing that a higher ratio of inliers is achieved  in the initial established feature correspondences  between two point clouds by matching NN-fused RCS features and RANSAC can quickly find the main cluster formed by inliers. For SHOT+RoPS+RCS fusion, we can see that NN outperforms all others with a significant margin. These results demonstrate that NN is able to generate distinctive representations during feature fusion on noisy and low-resolution Kinect data.  Finally, we present a qualitative result of two sample registration outcomes by tested methods in Fig.~\ref{fig:reg_view}. The figure suggests that the point clouds in the BMR datasets are very noisy and with limited geometric information. However, NN-based registration still accurately aligns the point cloud pairs while PCA and min pooling produce larger errors.

\paragraph{\textbf{2) Results on the Aug\_ICL-NUIM dataset}}
\begin{table}[t]\small
	\caption{Evaluation on the Aug\_ICL-NUIM benchmark. 3DMatch and CGF are deep learned features. ``Comb. 1'' and ``Comb. 2'' are the fusion results of RCS (48 dim.) and SHOT+RoPS+RCS (256 dim.), respectively. The best and second best results are shown in bold and underlined fonts, respectively.}
	\renewcommand{\arraystretch}{1}
	\centering
	\begin{tabular}{@{\extracolsep{\fill}}cccc}
		\hline
		&Precision (\%)&Recall (\%)&F-score (\%)\\
		\hline
		OpenCV~\cite{drost2010model}&1.6&5.3&2.5\\
		Super 4PCS~\cite{mellado2014super}&10.4&17.8&13.1\\
		PCL~\cite{rusu2009fast}&14.0&44.9&21.3\\
		FGR~\cite{zhou2016fast}&23.2&51.1&31.9\\
		CZK~\cite{choi2015robust}&19.6&59.2&29.4\\
		3DMatch~\cite{zeng20173dmatch}&\underline{25.2}&65.1&\underline{36.3}\\
		CGF (FGR)~\cite{khoury2017learning} &9.4&60.7&16.3\\
		CGF (CZK)~\cite{khoury2017learning}&14.6&\bf72.0&24.3\\
		Ours (Comb. 1)&15.8&58.6&24.9\\
		Ours (Comb. 2)&\bf26.0&\underline{71.2}&\bf38.1\\
		\hline
	\end{tabular} 
	
	\label{tab:scene_reg}
\end{table}
The results of our approach with 1000 RANSAC iterations  as well as benchmarking results of many state-of-the-art methods are reported in Table~\ref{tab:scene_reg}.

One can see that the fusion of RCS achieves comparable performance with CGF (CZK), while the fusion of SHOT+RoPS+RCS exceeds all compared methods. This result has demonstrated the effectiveness of our fusion approach from two aspects. First, the fusion of traditional local geometric descriptors can enhance their discriminative power greatly. Second, our method is not only more effective than existing fusion approaches, but also shows competitive performance to deep learned descriptors (e.g., 3DMatch~\cite{zeng20173dmatch} and CGF~\cite{khoury2017learning}). In addition, an appealing trait of our fused descriptor is intrinsically to rotation, as opposed to most of existing learned ones. We believe
even better results could be achieved by trying more distinctive feature sets. Although the selection of features for fusion is somewhat tricky, simply fusing RCS achieves better F-score performance than CGF (CZK). \textit{It is therefore not a cumbersome work for the selection of input features}. 

\begin{figure}[t]
	\centering
	\includegraphics[width=0.8\linewidth]{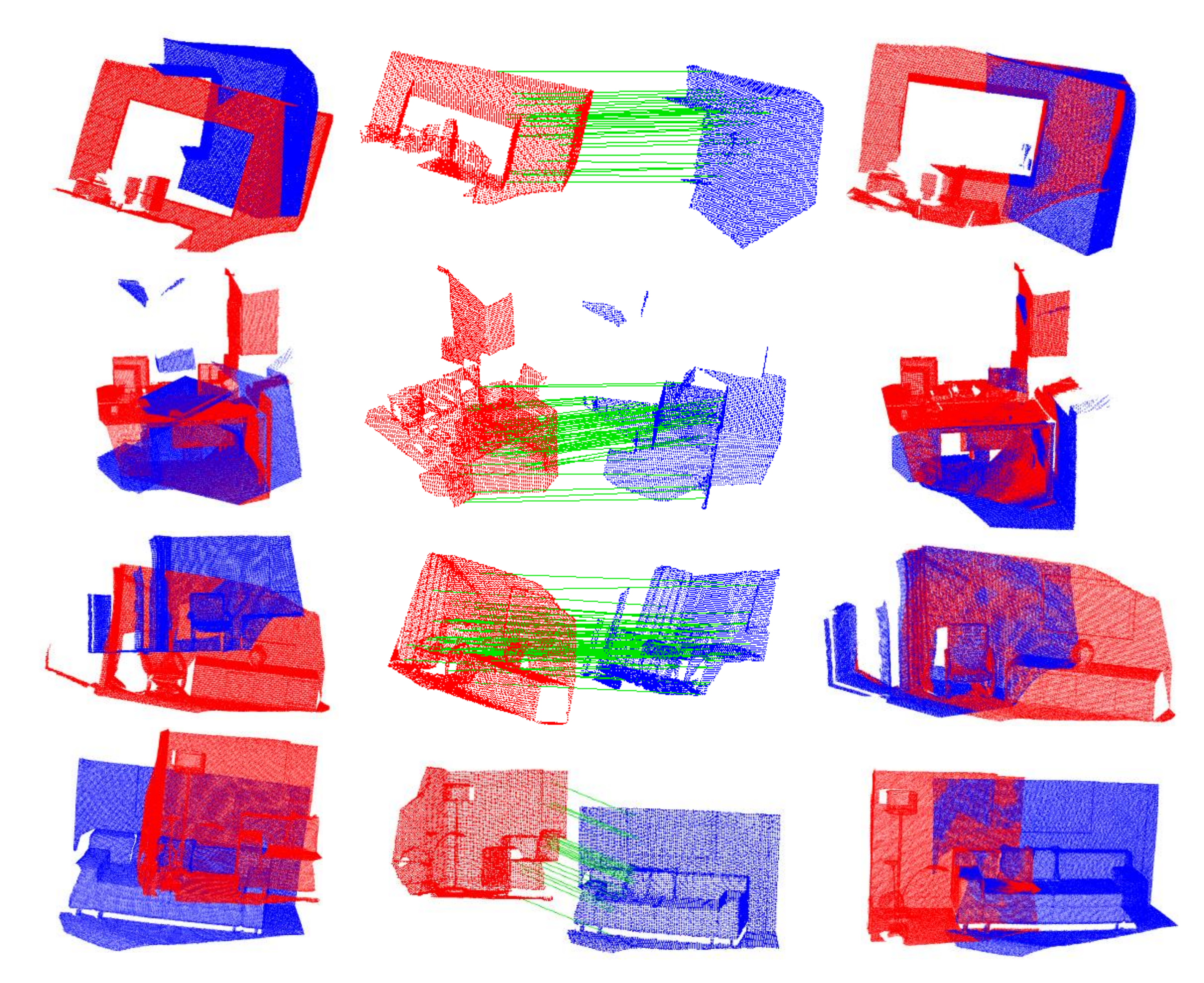}\\
	\caption{Visualization of four exemplar rigid geometric registration results from the Aug\_ICL-NUIM dataset based on the fusion descriptor of SHOT+RoPS+RCS by the proposed approach. From left to right: initial point clouds, point-to-point correspondences (green lines), and the aligned result.}
	\label{fig:scene_reg_view}
\end{figure}

Finally, we present some registration results on this dataset by the fusion descriptor of SHOT+RoPS+RCS in Fig.~\ref{fig:scene_reg_view}. Even for rigid data with limited geometric information, our method still achieves accurate registrations.
\section{Conclusions and future work}\label{sec:conc}
We have presented a simple yet effective  deep learning-based method for local geometric feature fusion that employs a neural network to mine complementary information within a set of geometric features. Feature matching and geometric registration experiments on several public available datasets with different modalities and application scenarios confirm that our method is able to achieve more distinctive features than existing fusion approaches while occupying less dimensions. The fused features are also competitive to several deep learned geometric descriptors yet being more lightweight to train and rotational invariant. In light of the experimental results, the following conclusions can be drawn.

\begin{itemize}
	\item In terms of feature fusion, the proposed fusion approach attains a comprehensive exploitation of the complementary information within a feature set using very compact representations.
	\item Besides superior fusion performance, the resultant descriptor after fusion achieves competitive performance to learned descriptors. Since one of the most challenging problems for learned point cloud features is how to achieve rotation invariance~\cite{yew20183dfeat}, our work directs a new way by fusing traditional features using a lightweight NN that improves their distinctiveness greatly while being rotational invariant.
\end{itemize}

Potential future research directions include (i) investigating more advanced training strategies (e.g., hard negative mining~\cite{simo2015discriminative,Lin2017Focal}) to further improve the fusion performance; (ii) the application our method to other domains such as 2D image feature and non-rigid feature description as the inference of our method can be seamlessly applied to these tasks.
\section*{Acknowledgments}
The authors would like to thank the publishers of experimental datasets used in our work. Financial supports from the National Natural Science Foundation of China under Grant 61876211  are gratefully acknowledged.
\section*{References}
\bibliography{mybibfile}

\end{document}